\documentclass{article}

\usepackage[preprint]{neurips_2021}

\usepackage[utf8]{inputenc} 
\usepackage[T1]{fontenc}    
\usepackage{hyperref}       
\usepackage{url}            
\usepackage{booktabs}       
\usepackage{amsfonts}       
\usepackage{nicefrac}       
\usepackage{microtype}      
\usepackage{xcolor}         

\usepackage{graphicx}
\usepackage{caption}
\usepackage{subcaption}
\usepackage{float}
\usepackage{amsmath}
\usepackage[normalem]{ulem}
\usepackage[linesnumbered,ruled,vlined]{algorithm2e}
\usepackage{authblk}
\usepackage{CJKutf8} 

\title{Learning Multilingual Representation for Natural Language Understanding with Enhanced Cross-Lingual Supervision}

\author[1]{Yinpeng Guo}
\author[1]{Liangyou Li}
\author[1]{Xin Jiang}
\author[1]{Qun Liu}
\affil[1]{Huawei Noah's Ark Lab\authorcr
  \{\tt yinpeng.guo, liliangyou, jiang.xin, qun.liu\}@huawei.com}

\begin{document}

\maketitle

\begin{abstract}
Recently, pre-training multilingual language models has shown great potential in learning multilingual representation, a crucial topic of natural language processing. 
Prior works generally use a single mixed attention (MA) module, following TLM \citep{CONNEAU_xlm_NEURIPS2019}, for attending to intra-lingual and cross-lingual contexts equivalently and simultaneously.
In this paper, we propose a network named decomposed attention (DA) as a replacement of MA.
The DA consists of an intra-lingual attention (IA) and a cross-lingual attention (CA), which model intra-lingual and cross-lingual supervisions respectively.
In addition, we introduce a language-adaptive re-weighting strategy during training to further boost the model's performance.
Experiments on various cross-lingual natural language understanding (NLU) tasks show that the proposed architecture and learning strategy significantly improve the model's cross-lingual transferability.
\end{abstract}

\section{Introduction}
Representation learning has become an important research area in natural language processing (NLP) for a long time.
It aims to build models to automatically discover features from raw text for downstream NLP tasks.
Beyond representation learning for specific languages, multilingual representation learning is recently drawing more attention.
It learns universal representation across languages and strongly improves performance of cross-lingual natural language understanding (NLU) tasks, especially zero-shot transfer on low-resource languages. 

Although overall performance of cross-lingual transfer has been remarkably improved by previous works, most models are built upon mixed attention (MA) modules. 
As shown in Figure~\ref{fig:struct_ma}, in MA modules, intra-lingual context and cross-lingual context are attended without distinction. 

Then, questions come out naturally: Why should intra-lingual context and cross-lingual context be mixed?
Is the model's capacity sufficient to capture intra-lingual and cross-lingual with a set of shared parameters? Is the model prefer intra-lingual context for word prediction?

In this paper, we propose decomposed attention (DA) which consists of an intra-lingual attention (IA) to learn from intra-lingual context and a following cross-lingual attention(CA) which depends on cross-lingual context.
Figure~\ref{fig:struct_ila-cla} shows the architecture of DA. This kind of independent modeling improves model capacity and is helpful to capture cross-lingual information, as in neural machine translation~\citep{Vaswani_NIPS2017_transformer}.
Figure~\ref{fig:attn} shows the attention heatmaps obtained from MA and DA, respectively. 
Comparing attentions between IA and CA in Figure~\ref{fig:attn_ma}, we find that MA prefers to distribute attentions on intra-lingual context than cross-lingual context.
In other words, MA predict words depending more on intra-lingual context. 
On the contrary, as shown in Figure~\ref{fig:attn_da}, DA utilizes cross-lingual context much better.

In addition to the above improvement in architectures, we further explore language-adaptive sample re-weighting  which encourages the model to learn challenging examples for each language. 
The difficulty of an example is defined as its prediction loss. 
In our experiment, the re-weighting strategy is language-wise and only triggered when easy examples are well learned.

In general, the contributions of this paper include:
\begin{itemize}
    \item We propose a DA network for multilingual representation learning, which consists of an intra-lingual attention and a cross-lingual attention.
    \item We propose a language-adaptive sample re-weighting strategy, which encourages the model to gradually learn difficult examples in language-wise.
    \item Experimental results on different types of multilingual NLU tasks and further analysis prove the effectiveness of the proposed DA network and re-weighting strategy.
\end{itemize}

\section{Related Work}
\paragraph*{Encoder-Only Models} In recent years, with great progresses in language model pre-training, multilingual representation learning methods based on pre-training are emerging in large numbers. 
Inspired by masked language modeling (MLM)~\citep{devlin-etal-2019-bert}, mBERT proposed multilingual MLM (mMLM) to learn representation from a mixture of Wikipedia corpora from one hundred languages. 
XLM~\citep{CONNEAU_xlm_NEURIPS2019} introduced parallel translation corpora to further incorporate a cross-lingual pre-training objective TLM with mMLM. 
TLM concatenates a pair of translation sentences as a single input sequence to Transformer encoder, then performs the masked word prediction objective as MLM.
Subsequently, following mBERT and XLM, various multilingual pre-training models were proposed.
XLM-Roberta~\citep{conneau-etal-2020-unsupervised} extended XLM via enlarging model size and training corpora.
Unicoder~\citep{huang-etal-2019-unicoder} introduced three additional cross-lingual pre-training tasks, i.e. cross-lingual word recovery, cross-lingual paraphrase identification and cross-lingual masked language model.
ERNIE-M~\citep{ouyang2021erniem} leveraged cross-lingual alignments to augment multilingual training data from monolingual corpora.
HICTL~\citep{wei2021on} designed a hierarchical contrastive learning framework on parallel corpora, including word-level and sentence-level contrast.

\paragraph*{Encoder-Decoder Models} In addition to encoder-only models mentioned above, encoder-decoder models were also explored in multilingual representation learning. 
Multilingual BART (mBART)~\citep{liu2020multilingual} and Multilingual T5 (mT5)~\citep{xue2020mt5} applied denoising auto-encoder strategies in multilingual pre-training.
VECO~\citep{luo2021veco} integrated MLM, TLM and translation pre-training and shared self-attention parameters of encoder and decoder.

\paragraph*{Decoder-Only Models} Besides, \citet{DBLP:journals/corr/abs-1911-03597} leveraged decoder-only Transformer network to learn multilingual language model for zero-shot paraphrase generation.

However, most of these models adopt MA as their encoder's default backbone.

\section{Methodology}

We adopt a Transformer encoder as the backbone of our model since we aim to tackle NLU tasks where decoder is unnecessary.
In addition, because the proposed DA module is an alternative to the conventional MA module, which is the default backbone of current SOTA encoder-decoder models, it is feasible to apply the DA module in encoder-decoder models.

\subsection{Multilingual Masked Language Modeling (MMLM)}
Since our model is based on Transformer encoder, it is convenient to follow the multilingual masked language modeling (MMLM) objective proposed by mBERT.
During training on monolingual corpora (e.g. Wikipedia), sentences from the same document are concatenated as a single sequence to fulfill the maximum input length of the encoder. 
Data from different languages are mixed so that model parameters are trained for various languages.
Before inputted into the model, words in the sequence are randomly replaced with probability of 15\%, in which 80\% is a [MASK] token, 10\% is a random word and the rest remain unchanged.
The training objective is to predict the masked words depending on other visible words in the same language as described in Figure~\ref{fig:struct_mmlm}, Appendix~\ref{app:mmlm}.
More specifically, the cross-entropy objective function of MMLM can be written as Equation~\ref{eq:loss_mmlm}:
\begin{equation}
    \label{eq:loss_mmlm}
    \begin{aligned}
    &\mathcal{L}_{MMLM} = -\sum_{x\in{D_m}}\sum_{i\in[1,n]}{\log{p(x_i|X/x_i)}}
    \end{aligned}
\end{equation}
where $D_m$ is the monolingual corpus, $n$ is the maximum sequence length, $X/x_i$ denotes all the input words in $X$ except for $x_i$ and $p(\cdot|\cdot)$ is the prediction probability.

In experiments, we apply MMLM objective on both baseline and our models.

\subsection{Decomposed Attention (DA)}
\begin{figure*}[ht]
	\centering
	\center 
	\begin{subfigure}[t]{.49\textwidth}
		\centering
		\includegraphics[width=1.\linewidth]{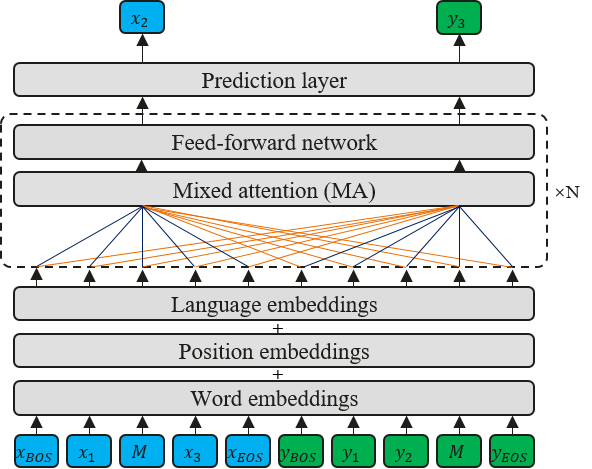}
		\caption{Mixed attention (MA).}
		\label{fig:struct_ma}
	\end{subfigure}
	\begin{subfigure}[t]{.49\textwidth}
		\centering
		\includegraphics[width=1.\linewidth]{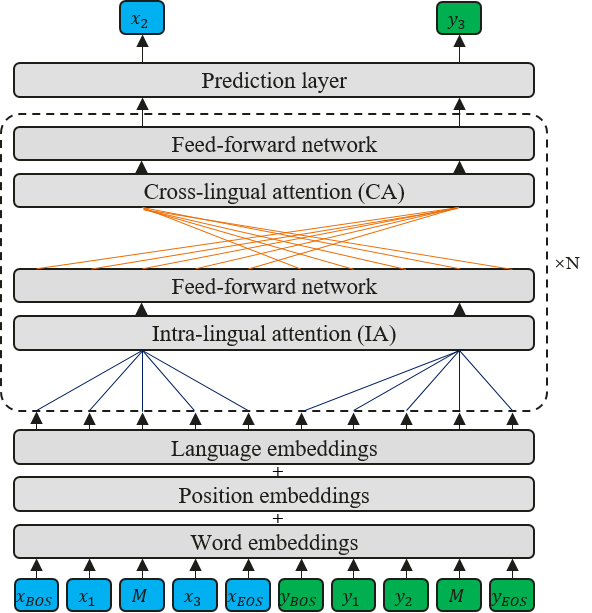}
		\caption{Decomposed attention (DA).}
		\label{fig:struct_ila-cla}
	\end{subfigure}
	\caption{Comparison between mixed attention (MA) and the proposed decomposed attention (DA): intra-lingual attention (IA) and cross-lingual attention (CA).}
	\label{fig:struct}
\end{figure*}

During training on cross-lingual parallel corpora, the input is a pair of masked translation sentences in two languages.
For instance in Figure~\ref{fig:struct}, the sentence $X=\{x_1, x_2, ..., x_n\}$ and its translation $Y=\{y_1, y_2, ..., y_n\}$ are the inputs.
The objective function of MA and DA are identical, which aims to predict masked words conditioning on both intra-lingual and cross-lingual context, except for the masked word itself, as described in Equation~\ref{eq:loss_trans}:
\begin{equation}
\label{eq:loss_trans}
\begin{aligned}
\mathcal{L}_{Trans} = -\sum_{(x,y)\in{D_b}}\Big[&\sum_{i=1}^{\frac{n}{2}}{\log{p(x_i|C_{x_i})}} + \sum_{j=1}^{\frac{n}{2}}{\log{p(y_j|C_{y_j})}}\Big] \\
\end{aligned}
\end{equation}
Where $D_b$ is the bilingual translation corpus, $C_{x_i}=(X\cup{Y})/x_i$ is the context of $x_i$ and $C_{y_j}=(X\cup{Y})/y_j$ is the context of $y_j$.

Although MA and DA share the same objective function, calculation details of their prediction probabilities are different.
Since MA and DA are built on attention mechanism, we can formalize the computation of attention as Equation~\ref{eq:attn}, following the notation from~\citet{Vaswani_NIPS2017_transformer},
\begin{equation}
    \label{eq:attn}
    \begin{aligned}
        \mathrm{Attention}(Q,K,V) &= \mathrm{softmax}(\frac{QK^T}{\sqrt{d_k}})V
    \end{aligned}
\end{equation}
Where $Q\in\mathbb{R}^{n\times{d_k}},K\in\mathbb{R}^{n\times{d_k}},V\in\mathbb{R}^{n\times{d_k}}$ are the query, key and value matrices respectively and $\sqrt{d_k}$ is the dimension size.
For simplicity, we use $\odot(\cdot)$ to represent common computations between MA and DA in following equations, including concatenation of multi-head attentions, layer normalization and linear projections.
Then prediction probability from MA can be illustrated as Equation~\ref{eq:pred_prob_ma},
\begin{equation}
    \label{eq:pred_prob_ma}
    \begin{aligned}
    &p(x_i|C_{x_i})_{MA}=\mathrm{softmax}(\odot(\mathrm{Attention}(Q_{x_i}W_Q,K_{C_{x_i}}W_K,V_{C_{x_i}}W_V)))
    \end{aligned}
\end{equation}
Where $W_Q,W_K,W_V\in\mathbb{R}^{d_{model}\times{d_k}}$ are projection matrices for a model with $d_{model}$-dimensions hidden states.
Apparently, in MA, cross-lingual and intra-lingual contexts are used together without distinction.

By contrast, the proposed DA module calculates CA based on the outputs of IA, as in Equation~\ref{eq:pred_prob_da}, 
\begin{align}
    \label{eq:pred_prob_da}
    p(x_i|C_{x_i})_{DA}&=\mathrm{softmax}(\odot(\mathrm{Attention}(Q_{x_i}^{CA},K_{Y}^{CA},V_{Y}^{CA}))) \\
    \mathrm{where}\;
    &Q_{x_i}^{CA}=\odot(\mathrm{Attention}(Q_{x_i}^{IA}W_{Q}^{IA},K_{X/x_i}^{IA}W_{K}^{IA},V_{X/x_i}^{IA}W_{V}^{IA})) \nonumber\\
    K_{Y}^{CA},&V_{Y}^{CA}=\odot(\mathrm{Attention}(Q_{Y}^{IA}W_{Q}^{IA},K_{Y}^{IA}W_{K}^{IA},V_{Y}^{IA}W_{V}^{IA})) \nonumber
\end{align}
$p(y_j|C_{y_j})_{MA}$ and $p(y_j|C_{y_j})_{DA}$ can be derived in the similar way. Note that the calculation of CA does not require projection matrices for reducing model parameters.

According to Equation~\ref{eq:pred_prob_da}, DA is able to model intra-lingual and cross-lingual dependencies separately.
This feature strengthens the model's capability of learning cross-lingual supervision.
Attention heatmap in Figure~\ref{fig:attn_da} also reveals that DA captures stronger cross-lingual context information than MA.

\subsection{Language-Adaptive Sample Re-weighting}\label{sec:reweighting}
In addition to concerns on network architecture, data imbalance is another issue that limits the model's multilinguality.
Masked language modeling is a word-level multi-classification problem which consists of vocabulary size $\mathcal{V}$ classes.
Due to the imbalance nature of word distribution, training examples of each class is extremely imbalance.
For example the sentence \textit{"The cat went away."}, \textit{"The"} and \textit{"."} appear 29.20M and 1009.06M times in our Wikipedia corpus, while \textit{"cat"}, \textit{"went"}, \textit{"away"} appear 0.61M, 0.67M, 0.42M times respectively.
In this example, the maximum imbalance ratio (IR) is 2403. 
However, the key semantic information is usually given by less frequent words, e.g. \textit{"cat went away"} in the example.
It is beneficial for the model to focus more on the less frequent words.

Hence, we introduce focal loss~\citep[FL][]{Lin_2017_ICCV} to address the class imbalance issue.
It was proposed to tackle class imbalance problem in objective detection.
As mentioned above, we formalize the masked language modeling as a word-wise multi-classification task.
Then, the Equations~\ref{eq:loss_mmlm} and \ref{eq:loss_trans} can be modified into
\begin{align}
    &\mathcal{L}_{\eta} = \alpha(1-p_t)^{\gamma}\mathcal{L}_{\eta} \label{eq:focal_loss}
\end{align}
Where $p_t = p(t|C_t)$, $t\in\{x_i, y_j\}$, $\eta\in\{MMLM,Trans\}$, $\alpha$ is a hyper-parameter to scale the loss and $\gamma$ is a hyper-parameter to balance the strength of focus on low-frequent and difficult examples.
The higher $\gamma$ is, the more focus on difficult examples.
In our experiments, we set $\alpha=0.25$ and $\gamma=2$ as suggested by~\citet{Lin_2017_ICCV}.
To further understand FL, we refer readers to its original paper~\citep{Lin_2017_ICCV}.

According to Equation~\ref{eq:focal_loss}, FL encourages the model to focus more on words of lower prediction probability (higher CE loss).
In another word, FL pushes the model to learn more from difficult examples.
But in our preliminary experiments, we found that the learning difficulty (loss) is varied among languages or between data types, as depicted in Figure~\ref{fig:lang_loss}, Appendix~\ref{app:lang_loss}.
We also observe that naive FL works not stably if directly applied with different initializations~\footnote{This is also addressed by initialization tricks in~\citet{Lin_2017_ICCV}, and also discussed and explained by~\citet{fl_leimao}. 
In brief, with random or inappropriate initialization, FL would also weight the high-frequent words since they are not learned well. 
However, what we want is to only weight low-frequent words.}.
Therefore, applying FL on all training samples at the same time will restrain the learning of low-frequent and difficult samples, such as those from low-resource languages and monolingual corpora.

On basis of above analysis, we design a language-wise and data-type-wise FL to perform language-adaptive sample re-weighting.
Briefly speaking, we only apply FL to languages and data types which have been learned well.
Then, Equation~\ref{eq:focal_loss} is modified into
\begin{align}
    \mathcal{L}_{\eta} &= \alpha(1-p_t)^{\gamma_{l}}\mathcal{L}_{\eta} \label{eq:la-reweight} \\
        \gamma_{l} &=
        \begin{cases}
        0 &\text{, $\mathcal{L}_{CE}>\mathcal{L}_{T}$ or $step<S$} \nonumber \\
        \gamma &\text{, $\mathcal{L}_{CE}\le\mathcal{L}_{T}$ and $step\ge{S}$} \nonumber
        \end{cases}
\end{align}
Where $\gamma_{l}$ is a language-wise FL hyper-parameter, $\mathcal{L}_{T}$ and $S$ denote threshold value of training loss and training step respectively.
$\gamma_{l}$ is updated during training for different languages and data types independently.
Finally, the objective function of our model becomes
\begin{align}
    \label{eq:total_loss}
    \mathcal{L}=\sum_{l}\sum_{\eta}{\alpha(1-p_t)^{\gamma_{l}}\mathcal{L}_{\eta}}
\end{align}

\section{Experiments}

\subsection{Dataset and Preprocessing}
\subsubsection{Pre-Training}\label{sec:data_pt}
The 12-layer (BASE) models are pre-trained on Wikipedia dumps~\footnote{https://meta.wikimedia.org/wiki/List\_of\_Wikipedias} and OpenSubtitles-2018~\citep{lison-etal-2018-opensubtitles2018} corpus, which are under CC-BY-SA and BSD-licensed respectively.
We collected 59 languages of Wikipedia dumps (20200520 version).
We adopted opus-tools~\footnote{https://pypi.org/project/opustools-pkg/} to fetch 422 language pairs of OpenSubtitles-2018 translation corpus covering languages in our Wikipedia data.
Detailed statistics of our data are given in Appendix~\ref{app:dataset}.
We extracted pure text from Wikipedia and OpenSubtitles using WikiExtractor~\footnote{https://github.com/attardi/wikiextractor} and opus-tools respectively.
Byte-level Byte-Pair Encoding (BBPE)~\citep{wei2021training} was applied for tokenization. 
The vocabulary was created based on our Wikipedia data, consisting of 100618 tokens.

The 6-layer (MINI) models are pre-trained with 0.1\% Wikipedia data and 0.01\% OpenSubtitles-2018 data. The number of examples of each language pair is greater than one thousand. 
Other data configuration follows that of the BASE models.

\subsubsection{Fine-Tuning}
\textbf{XNLI}~\citep{conneau-etal-2018-xnli} is a multilingual version of the natural language inference (NLI) task which is translated from MultiNLI~\citep{williams-etal-2018-broad}. 

\noindent\textbf{PAWS-X}~\citep{yang-etal-2019-paws} is a multilingual version of paraphrase identification task which is translated from PAWS-X~\citep{zhang-etal-2019-paws}. 

\noindent\textbf{TyDiQA-GoldP} is a SQuAD-like multilingual question answering task that derived from ~\citet{clark-etal-2020-tydi}. Given a context passage and a question, the task is to predict a answer span from the passage. It covers 9 typologically diverse languages. 

\noindent\textbf{UDPOS}~\citep{nivre-etal-2020-universal} is a POS tagging task which covers 90 languages.
In this paper, we follow the setting of \citet{pmlr-v119-hu20b} that includes the 29 languages with validation data.

\subsection{Training Details}
\label{sec:train_details}
For fair comparison, we train mBERT, MA and the proposed DA models on identical corpora.
The motivation of this work is to explore a novel framework to leverage cross-lingual supervision better than the MA module.
Most SOTA multilingual models are based on MA and this work is orthogonal to them.
Hence, we only reproduce mBERT and MA as our baselines.
Experiments are conducted on Nvidia V100 GPUs.
Detailed hyper-parameter configuration is given in Appendix~\ref{app:hyper}.

\subsubsection{Pre-Training}
\textbf{mBERT} is trained only on Wikipedia corpus with masked word prediction objective and next sentence prediction (NSP) objective. 
Language embeddings are not adopted.

\textbf{MA} and \textbf{DA}-series models are trained on a mixture of Wikipedia and OpenSubtitles-2018 data. 
Both of them are only trained with masked word prediction objective. 
Batches of Wikipedia and OpenSubtitles-2018 examples are iteratively fed to the model. For each batch, only one type of examples is included. 
Language embeddings are trained with these models.

\subsubsection{Fine-Tuning}
All the evaluation of downstream NLU tasks are performed under the setting of zero-shot cross-lingual transfer.
In more details, we only fine-tune the models using English training and development set.
Then directly evaluate the models on testing set of the target language.

We adopt early stopping as the training stop criteria for all experiments.
The training is stopped if without accuracy increase on development set for 50 epochs.
In order to reduce randomness, the results are reported as the average score of 8 rounds of experiments with different random seeds.

For sentence pair classification, i.e. \textbf{XNLI} and \textbf{PAWS-X}, two sentences are concatenated as a single sequence before fed into the models.
Then the contextual word representation of the [BOS] is used by downstream classification network. 
All of the models adopt identical classification network, following the original mBERT.

For question answering and structured prediction tasks, i.e. \textbf{TyDiQA-GoldP} and \textbf{UDPOS}, contextual representations of all words at the last layer are extracted as the input to downstream network.
In TyDiQA-GoldP, a pair of question and its related answering passage is concatenated as models' input.
The model is trained to label the answer span's start word and end word.
Similarly, model is trained to label POS of each input word in UDPOS task.
DA models only adopt IA layers for question answering and structured prediction tasks.

Fine-tuning hyper-parameters of all models are identical. 
Detailed configuration is provided in Appendix~\ref{app:hyper}.

\subsection{Experimental Results}
\label{sec:exp}
In this section, \textbf{+ FL} and \textbf{+ Adapt-FL} denote DA models trained with naive focal loss and the proposed language-adaptive sample re-weighting respectively.
We adopt language code from ISO-639-1 as abbreviation of the tested languages.

\subsubsection{Sentence Pair Classification}

\begin{table*}[!ht]
    \centering
    \resizebox{1.\textwidth}{9.8mm}{
    \begin{tabular}{l|ccccccccccccccc|c}
        \hline 
        Model & en & ar & bg & de & el & es & fr & hi & ru & sw & th & tr & ur & vi & zh & avg \\
        \hline 
        mBERT & 80.2 & 64.3 & 69.7 & 69.2 & 67.2 & 72.9 & 71.6 & 58.7 & 68.4 & \textbf{55.6} & 49.8 & 62.2 & 58.1 & 69.1 & 65.4 & 65.5  \\
        MA & 81.7 & 70.4 & 75.9 & 73.9 & 73.6 & 77.0 & 76.1 & 61.5 & 73.3 & 47.9 & 68.8 & 71.6 & \textbf{60.7} & 73.6 & 68.8 & 70.3 \\
        \hline 
        DA & \textbf{82.0} & \textbf{71.2} & 76.3 & 74.7 & 73.9 & 77.4 & \textbf{76.8} & 61.8 & \textbf{73.5} & 46.6 & 68.9 & 71.3 & 59.6 & \textbf{74.5} & 69.2 & 70.5 \\
        \hspace*{0.5em}+ FL & 81.7 & 71.1 & \textbf{76.4} & 74.5 & \textbf{74.2} & 77.3 & 76.4 & 62.5 & 73.3 & 48.2 & \textbf{69.3} & 71.5 & 60.0 & 73.9 & 69.2 & 70.6 \\
        \hspace*{0.5em}+ Adapt-FL & 81.8 & 70.9 & 76.3 & \textbf{75.0} & 74.1 & \textbf{77.6} & 76.6 & \textbf{62.9} & 73.3 & 48.1 & 69.0 & \textbf{71.7} & 60.2 & 74.4 & \textbf{69.3} & \textbf{70.8} \\
        \hline 
    \end{tabular}
    }
    \caption{XNLI accuracy scores.} 
    \label{tab:base-xnli}
\end{table*}

We first evaluate baselines and our models on the widely used multilingual classification task XNLI.
As shown in Table~\ref{tab:base-xnli}, DA models consistently perform better than MA on most languages and on average.
It demonstrates the effectiveness of the proposed DA module.
With the proposed sample re-weighting strategy, DA's performance is further improved.
Performance on relatively low-resource languages are consistently improved, e.g. Hindi (hi), Swahili (sw), Turkish (tr) and Urdu (ur).

We also notice that mBERT outperforms both DA and MA on Swahili.
It's because Swahili is not included in the OpenSubtitles-2018 translation corpus, but included in the Wikipedia monolingual corpus.
Since DA and MA models are iteratively trained over translation and monolingual corpora, Swahili is underfit by the two models.

\begin{table*}[!ht]
	\centering
	\resizebox{.7\textwidth}{10.4mm}{
		\begin{tabular}{l|ccccccc|c|cc}
			\hline 
		    Model & en & de & es & fr & ja & ko & zh & avg & euro & asia \\
			\hline 
			mBERT & 92.9 & 82.3 & 85.1 & 86.8 & 65.6 & 65.5 & 74.8 & 79.0 & 84.7 & 68.6 \\
			MA & 93.0 & 83.7 & 87.9 & 87.8 & 62.0 & 65.1 & 74.5 & 79.1 & 86.5 & 67.2 \\
			\hline 
			DA & 93.3 & 84.8 & 88.0 & 88.8 & 67.2 & \textbf{72.2} & 78.5 & 81.8 & 87.2 & 72.6 \\
			\hspace*{0.5em}+ FL & 93.4 & 85.2 & \textbf{88.5} & \textbf{88.9} & 66.9 & 72.1 & 78.9 & 82.0 & \textbf{87.5} & 72.6 \\
			\hspace*{0.5em}+ Adapt-FL & \textbf{93.6} & \textbf{85.3} & 88.2 & \textbf{88.9} & \textbf{68.1} & \textbf{72.2} & \textbf{79.0} & \textbf{82.2} & \textbf{87.5} & \textbf{73.1} \\
			\hline 
		\end{tabular}
	}
	\caption{PAWS-X accuracy scores. \textit{euro}: average score of German (de), Spanish (es) and French (fr). \textit{asia}: average score of Japanese (ja), Korean (ko) and Chinese (zh).} 
	\label{tab:base-pawsx}
\end{table*}

In order to diagnose DA's advantage on cross-lingual alignment, we evaluate DA and the baselines on a multilingual paraphrase identification task PAWS-X.
Better cross-lingual alignment is supposed to benefit paraphrase identification more than natural language inference, since word alignment in paraphrase pair is stricter.
As illustrated in Table~\ref{tab:base-pawsx}, DA outperforms the baselines with a large gap on most languages.

This advantage is more obvious on Asian languages than on European languages.
According to Table~\ref{tab:base-pawsx}, MA beats mBERT on \textit{euro} but is defeated on \textit{asia}.
It implies that given translation corpora, MA is only able to learn cross-lingual alignment from similar languages, such as the ones within the same language family of the source language.
However, DA is able to learn stronger alignment even between languages with different syntax.
We further analyze this feature in Section~\ref{sec:analysis_da}.
Besides, compared with naive FL, the improvement on \textit{ja, ko and zh} brought by the language-adaptive sample re-weighting demonstrates its effectiveness again, since sample frequency of \textit{aisa} is generally lower than \textit{euro} in the training set.

\subsubsection{Question Answering}

\begin{table*}[!ht]
    \centering
    \resizebox{1.\textwidth}{9.5mm}{
    \begin{tabular}{l|ccccccccc|c}
        \hline 
        Model & en & ar & bn & fi & id & ko & ru & sw & te & avg \\
        \hline 
        mBERT & 56.8 / \textbf{43.9} & 41.9 / \textbf{25.9} & 23.9 / 16.5 & 35.6 / 22.1 & 44.0 / 29.6 & 23.5 / \textbf{17.6} & 39.8 / 25.6 & \textbf{39.2} / \textbf{24.9} & 26.9 / 23.1 & 36.9 / 25.5  \\
        MA & 46.6 / 22.4 & 36.9 / 12.1 & 20.4 / 36.6 & 31.9 / 20.2 & 38.0 / 26.7 & 18.9 / 11.0 & 36.0 / 22.7 & 34.5 / 21.2 & 19.2 / 13.9 & 31.4 / 20.7 \\
		\hline 
        DA & 54.6 / 30.5 & 47.6 / 19.2 & 27.5 / 42.5 & 40.9 / 27.4 & 47.2 / 32.3 & 22.0 / 15.2 & 44.9 / 30.5 & 37.6 / 23.2 & 30.0 / 23.7 & 39.2 / 27.2 \\
        \hspace*{0.5em}+ FL & \textbf{58.6} / 32.2 & \textbf{50.3} / 19.0 & \textbf{28.9} / \textbf{45.4} & 42.6 / 26.7 & 49.7 / 33.7 & \textbf{23.6} / 16.0 & \textbf{47.4} / \textbf{33.1} & 36.1 / 21.2 & 28.4 / 22.5 & \textbf{40.6} / \textbf{27.8} \\
        \hspace*{0.5em}+ Adapt-FL & 57.0 / 31.5 & 49.0 / 16.7 & 26.0 / 44.6 & \textbf{42.8} / \textbf{28.1} & \textbf{49.9} / \textbf{34.8} & 23.2 / 16.7 & 46.7 / 32.1 & 36.5 / 21.8 & \textbf{30.7} / \textbf{24.1} & 40.2 / \textbf{27.8} \\
        \hline 
    \end{tabular}
    }
    \caption{TyDiQA-GoldP F1 / EM scores.} 
    \label{tab:base-tydi}
\end{table*}

In addition to classification tasks, we evaluate the models on multilingual question answering.
Both F1 and exact match (EM) scores are reported in Table~\ref{tab:base-tydi}.
DA performs better than MA on all languages in terms of both F1 and EM scores.
In this task, MA performs worse even than mBERT, which can be also observed (as XLM) in~\citet{pmlr-v119-hu20b}.
We guess the reason for MA's failure is: 
question answering is a task requires long-distance dependency, the model needs to link the question span (at the beginning of the sequence) to the answer span (in the middle of the sequence). 
But sentence-level translation corpora (e.g. OpenSubtitles) usually consist of short sequences, which limits the model's long-distance modeling capability, then brings side-effects for question answering.
Especially for MA-based models that learn monolingual and translation corpora with mixed IA and CA.
However, DA decomposes IA and CA so that IA is not affected by short translation sequences.
Hence long-distance modeling capability is retained by DA's IA layers, then question answering can be addressed just as mBERT does.
Note that in the experiments, we only adopt DA's IA layers for the question answering task.
Besides, due to lack of Swahili translation data, MA and DA perform worse on Swahili again.

\subsubsection{Structured Prediction}

\begin{table*}[ht]
    \centering
    \resizebox{1.\textwidth}{20.5mm}{
    \begin{tabular}{l|ccccccccccccccccc}
        \hline 
              Model & en & af & ar & bg & de & el & es & et & eu & fa & fi & fr & he & hi & hu \\
        \hline 
        mBERT & 95.0 & 88.2 & 56.0 & 86.9 & 85.5 & 82.2 & 87.5 & 86.3 & 69.0 & 65.3 & 80.9 & 56.5 & 74.4 & 69.4 & 80.7 \\
        MA & 95.0 & 88.6 & {60.1} & {88.7} & 85.9 & 83.7 & {88.6} & 86.7 & 71.3 & {68.8} & {83.2} & 58.8 & 76.7 & 67.9 & 80.3 \\
		\hline 
        DA & 95.0 & {88.7} & 58.9 & 88.5 & {86.3} & {83.9} & 88.0 & {87.4} & 71.8 & 68.5 & 82.8 & 62.0 & 78.0 & 70.9 & {81.1} \\
        \hspace*{0.5em}+ FL & 95.1 & {88.7} & 58.2 & 88.4 & 86.1 & 83.3 & 87.8 & 87.3 & 71.6 & 67.8 & 82.8 & 61.5 & 77.7 & 71.7 & 80.8 \\
        \hspace*{0.5em}+ Adapt-FL & {95.2} & 88.6 & 59.6 & 88.6 & {86.3} & 83.6 & 88.4 & 87.1 & {71.9} & 68.1 & 82.6 & {65.1} & {78.3} & {72.2} & 81.0 \\
        \hline 
              & id & it & ja & ko & mr & nl & pt & ru & ta & te & tr & ur & vi & zh & \textbf{avg} \\
        \hline 
        mBERT & 81.8 & 85.4 & 58.9 & 49.6 & 62.1 & 88.3 & 87.9 & 87.0 & 41.9 & 65.8 & 66.7 & 62.1 & 54.9 & 63.3 & 73.1 \\
        MA & 82.2 & {87.1} & 61.9 & {54.8} & 61.4 & 89.4 & 88.5 & 88.1 & 62.4 & {67.0} & {73.5} & 60.7 & 56.2 & 65.5 & 75.3 \\
		\hline 
        DA & 82.5 & 86.3 & 61.9 & 53.1 & 61.8 & {89.5} & 88.5 & {88.7} & 64.0 & 65.1 & 71.6 & 62.6 & {56.9} & {68.6} & 75.6 \\ 
        \hspace*{0.5em}+ FL & {82.6} & 86.2 & {63.0} & 53.1 & 62.7 & {89.5} & 88.3 & 88.4 & {65.1} & {67.0} & 71.3 & 62.8 & 56.0 & 67.3 & 75.6 \\
        \hspace*{0.5em}+ Adapt-FL & 82.5 & 86.6 & 62.2 & 53.4 & {62.9} & {89.5} & {88.7} & 88.4 & 64.0 & 66.0 & 71.1 & {63.4} & 56.3 & 64.9 & \textbf{75.7} \\
        \hline 
    \end{tabular}
    }
    \caption{Accuracy scores on UDPOS validation set, covering 29 languages.} 
    \label{tab:base-pos}
\end{table*}

Finally, we evaluate the models on the multilingual POS task UDPOS.
Since DA only adopts IA layers during fine-tuning on UDPOS, the advantage of CA layers' cross-lingual modeling may not be fully leveraged.
Nevertheless, even with only IA layers, DA performs on par with MA as shown in Table~\ref{tab:base-pos}.

It's worth noting that both DA and MA are significantly better than mBERT, which implies cross-lingual supervision benefits the multilingual structured prediction task.
Hence, it is very promising to further explore how to fully take advantage of DA's network for structured prediction task.

\section{Further Analysis}
\subsection{Decomposed Attention}
\label{sec:analysis_da}

\begin{figure}[!ht]
	\centering
	\center 
	\begin{subfigure}[t]{.49\textwidth}
		\centering
		\includegraphics[width=1.\linewidth]{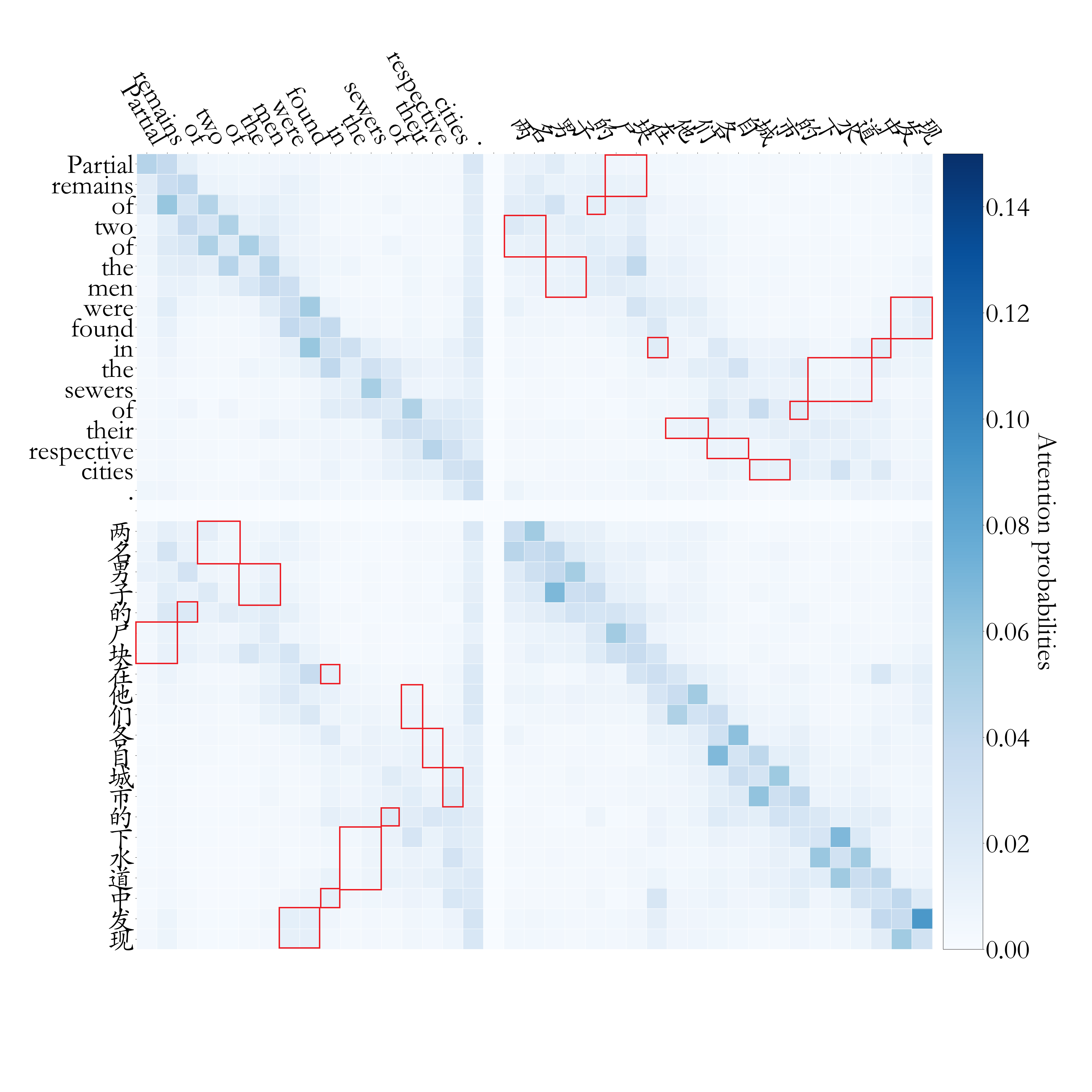}
		\caption{Mixed attention.}
		\label{fig:attn_ma}
	\end{subfigure}
	\begin{subfigure}[t]{.49\textwidth}
		\centering
		\includegraphics[width=1.\linewidth]{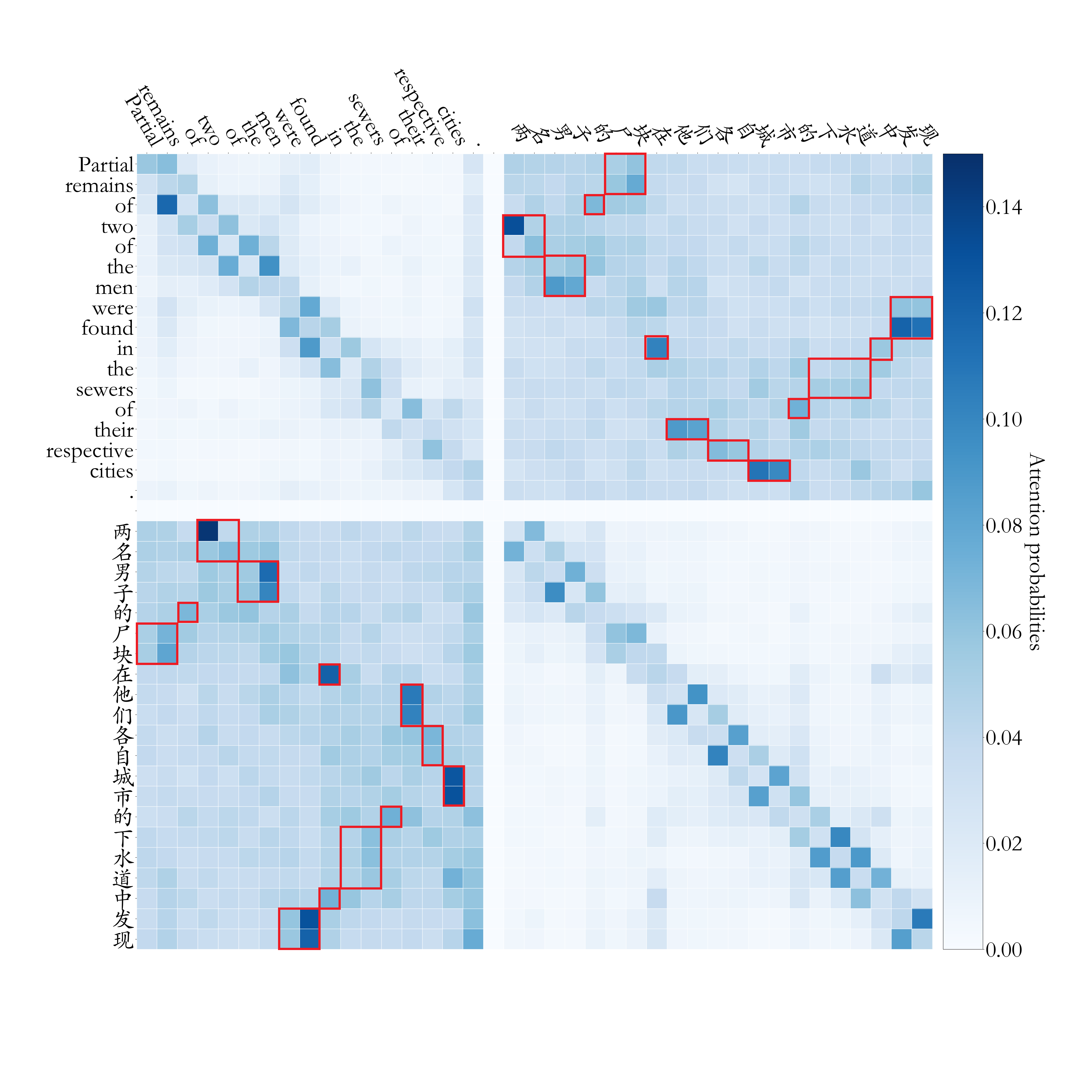}
		\caption{Decomposed attention.}
		\label{fig:attn_da}
	\end{subfigure}
	\caption[l]{Attention probability heatmap of MA and DA. Words attend from \textit{row} to \textit{column}. \textit{Upper-left}: IA of English. \textit{Lower-right}: IA of Chinese. \textit{Upper-right}: CA of English-Chinese. \textit{Lower-left}: CA of Chinese-English. \textit{{\color{red}Red blocks}}: reference alignments.}
	\label{fig:attn}
\end{figure}

Figure~\ref{fig:attn} provides a visual comparison between MA and DA in both intral-lingual attention (IA) and cross-lingual attention (CA).
Attention probabilities are colored at the same scale, $[0,0.15]$, as shown by the colorbars on the right side.
We observe DA's advantages as follows.

\noindent\textbf{Balance between IA and CA}
As shown in Figure~\ref{fig:attn_ma}, MA's attention probabilities on intra-lingual context (upper-left and lower-right) are obviously higher than those on cross-lingual context (upper-right and lower-left).
In contrast with MA, Figure~\ref{fig:attn_da} reveals significantly more balance attention between intra-lingual context and cross-lingual context, especially on the correctly aligned words.
The behaviour of MA is easy to understand since learning alignment from intra-lingual context is easier than from cross-lingual context.
However, because DA decomposes IA and CA, the model is forced to learn alignment from cross-lingual context.

\noindent\textbf{More accurate alignment}
Although MA is able to provide correct intra-lingual alignments, it's not as strong as DA.
Besides, intra-lingual alignments are monotonic which is so easy to model that cannot discriminate models' cross-lingual transferability.
Hence, we study the cross-lingual alignment of MA and DA in more details.
\begin{CJK}{UTF8}{gbsn}
The example in Figure~\ref{fig:attn} is challenging for CA, since its word order in English and Chinese are very different.
We labeled the reference alignment in Figure~\ref{fig:attn} in red blocks.
To understand Figure~\ref{fig:attn}, we define a correct alignment as the \textit{darkest} color block in a row which hits the golden alignment word.
Then from Figure~\ref{fig:attn_da}, we can observe that DA captures almost all the cross-lingual alignment accurately, in both English-Chinese and Chinese-English attention directions.
Even the long-distance prepositional phrase \textit{"在...中(in)"} is successfully aligned with high probabilities.
On the contrary, MA totally fails most cross-lingual alignments, such as \textit{"Partial remains"}, all \textit{"of"}s, \textit{"cities"} etc.

\noindent\textbf{Entity}
In more case studies, we find DA is especially powerful in capturing entities.
Examples are provided in Appendix~\ref{app:attn}.
Figure~\ref{fig:attn_entity} shows that DA highlights the name \textit{Maud Gonne}, and also successfully captures the alignment \textit{"'s (的) continuous (无数次) 拒绝 (refusal)"} while MA fails.

In conclusion, MA pay more attentions on intra-lingual context and cannot accurately capture cross-lingual alignment.
MA tends to simply provide monotonic cross-lingual alignment, like the intra-lingual alignment.
This issue may be due to its nature of mixed intra-lingual and cross-lingual attentions.
The proposed DA network tackles MA's defects and highly enhances model's capability in learning cross-lingual supervision.

\end{CJK}

\subsection{Effect of Model Size}
\label{sec:model_size}

\begin{table*}[!h]
    \begin{subtable}[h]{0.45\textwidth}
        \centering
        \resizebox{0.5\textwidth}{10mm}{
        \begin{tabular}{l|c}
            \hline 
            Model & \#Parameters\\
            \hline 
            \multicolumn{2}{l}{\textit{MINI models}} \\
            \hline 
            MA & 114.4 M \\
            \hline 
            DA & 117.8 M \\
            DA-reduce & 114.6 M \\
            DA-share & 114.4 M \\
            \hline 
        \end{tabular}
        }
        \caption{\textit{MINI}: 6 Transformer layers.}
        \label{tab:model-size_mini}
    \end{subtable}
    \hfill
    \begin{subtable}[h]{0.45\textwidth}
        \centering
        \resizebox{0.4\textwidth}{7mm}{
        \begin{tabular}{l|c}
            \hline 
            Model & \#Parameters\\
            \hline 
            \multicolumn{2}{l}{\textit{BASE models}} \\
            \hline 
            MA & 154.9 M \\
            \hline 
            DA & 161.7 M \\ 
            \hline 
        \end{tabular}
        }
        \caption{\textit{BASE}: 12 Transformer layers.}
        \label{tab:model-size_base}
    \end{subtable}
    \caption{Comparison of model size.}
    \label{tab:model-size}
\end{table*}

Although CA in the proposed model does not adopt attention projection matrices $W_{Q}^{CA},W_{K}^{CA},W_{V}^{CA}$, as shown in Equation~\ref{eq:pred_prob_da}, CA keeps parameters of the feedforward (FFN) layer and the subsequent layer normalization (LayerNorm).
It slightly improves number of DA's parameter.
Table~\ref{tab:model-size} illustrates the number of parameters of MA and DA models.
In order to verify the effect of model size and exactly align DA's model size to MA's, we propose two methods to reduce DA's model size:
1) \texttt{DA-reduce}: reduce the hidden dimension of Transformer layers from 768 to 720, but keep the dimension of embedding layer unchanged.
2) \texttt{DA-share}: share FFN and LayerNorm parameters between IA and CA.
Take MINI models for example, as shown in Table~\ref{tab:model-size_mini}, number of parameters of \texttt{DA-reduce} and \texttt{DA-share} are controlled identical to the MA counterpart.

\begin{table*}[!ht]
    \centering
    \resizebox{1.\textwidth}{8.5mm}{
    \begin{tabular}{l|ccccccccccccccc|c}
        \hline 
        Model & en & ar & bg & de & el & es & fr & hi & ru & sw & th & tr & ur & vi & zh & avg \\
        \hline 
        MA & 71.2 & 53.6 & 56.8 & 56.9 & 54.9 & 59.7 & 59.4 & 51.2 & 55.4 & 43.0 & 53.4 & 51.4 & 51.3 & 57.5 & 52.5 & 55.2  \\
        \hline 
        DA & 74.2 & 54.4 & 58.9 & 59.5 & 55.0 & 63.4 & 61.3 & 53.5 & 57.8 & 39.4 & 53.4 & 51.3 & 52.8 & 58.3 & 52.0 & 56.3 \\
        DA-reduce & 74.0 & 54.8 & 59.7 & 60.1 & 57.0 & 63.6 & 62.3 & 53.3 & 58.2 & 41.2 & 54.0 & 53.2 & 52.8 & 59.1 & 51.9 & 57.0 \\
        DA-share & 73.7 & 55.2 & 58.9 & 59.1 & 57.4 & 62.9 & 61.8 & 53.1 & 57.4 & 41.8 & 54.9 & 52.0 & 52.7 & 58.4 & 51.9 & 56.8 \\
        \hline 
    \end{tabular}
    }
    \caption{XNLI accuracy scores.} 
    \label{tab:mini-xnli-param}
\end{table*}

We further pre-train MINI models of MA, DA, DA-reduce and DA-share, then evaluate them on XNLI.
The MINI pre-training configuration is described in Section~\ref{sec:data_pt}.
Fine-tuning setting follows the BASE models, as introduced in Section~\ref{sec:train_details}.
The results in Table~\ref{tab:mini-xnli-param} demonstrate that the DA outperforms MA even with identical number of parameters.
\footnote{To our surprise, both \texttt{DA-reduce} and \texttt{DA-share} improves the performance of DA.
The results imply that:
1) linear projection between embeddings and Transformer hidden representations may enhance model capability, 720 hidden dimensions is large enough for MINI models.
2) the shared FFN and LayerNorm encourage IA and CA to build a language-agnostic semantic space.
We leave these conjectures for future research.}

\section{Conclusions and Future Directions}
In this paper, we explore the potential of making better use of cross-lingual supervision for learning multilingual representation in two aspects.
In terms of network architecture, we propose a decomposed attention (DA) module equipped with independent intra-lingual attention (IA) and cross-lingual attention (CA).
Compared to conventional, widely-used mixed attention (MA) network, DA significantly improves models' learning ability on cross-lingual supervision.
We further propose a language-adaptive sample re-weighting strategy for learning multilingual representation.
The cross-lingual transferability of the model is further boosted by the re-weighting strategy.
Experiments on various cross-lingual NLU tasks and further analysis prove the effectiveness of the proposed methods.

However, there are still questions required future research.
First, as mentioned in Section~\ref{sec:exp}, currently we only adopt DA's IA layers for structured prediction task.
There should be more effective ways to explore DA's potential in this task.
Another direction may be related to the re-weighting strategy.
Currently, the its hyper-parameters is empirically set.
We are curious if there is a theoretical method to decide it directly, without prior information of the training data.
There are many interesting topics on multilingual representation learning, we leave these topics for future works.

\bibliographystyle{acl_natbib}
\bibliography{ref_conf,ref_books,ref_arxiv,ref_journ,ref_urls}

\clearpage
\appendix

\section{Attention Heatmaps}
\label{app:attn}
In this section, we provide more attention examples to reveal the advantage of DA over MA.

\subsection{DA Highlights Entities}

\begin{figure}[!h]
	\centering
	\center 
	\begin{subfigure}[t]{.49\textwidth}
		\centering
		\includegraphics[width=1.\linewidth]{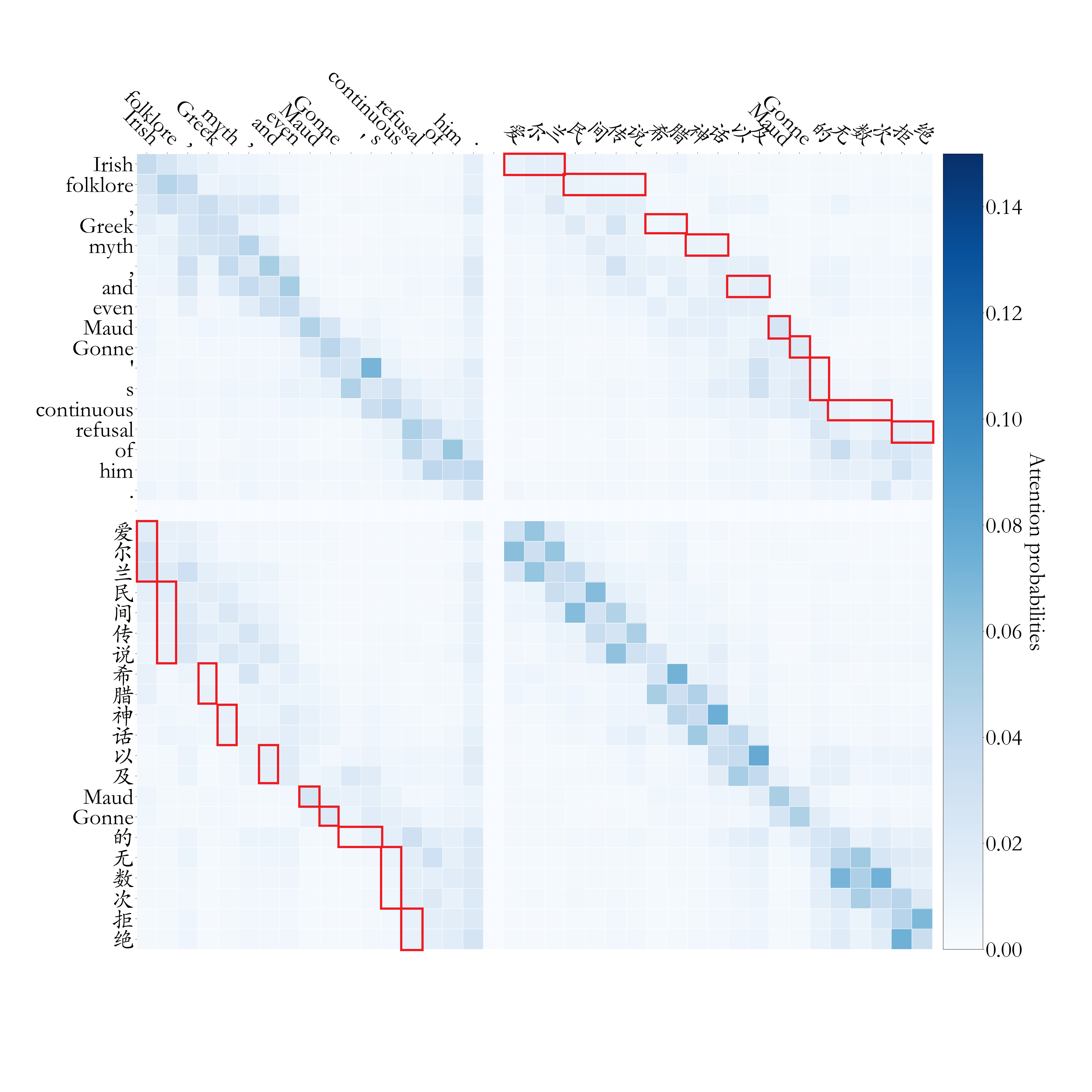}
		\caption{Mixed attention.}
		\label{fig:attn_ma_entity}
	\end{subfigure}
	\begin{subfigure}[t]{.49\textwidth}
		\centering
		\includegraphics[width=1.\linewidth]{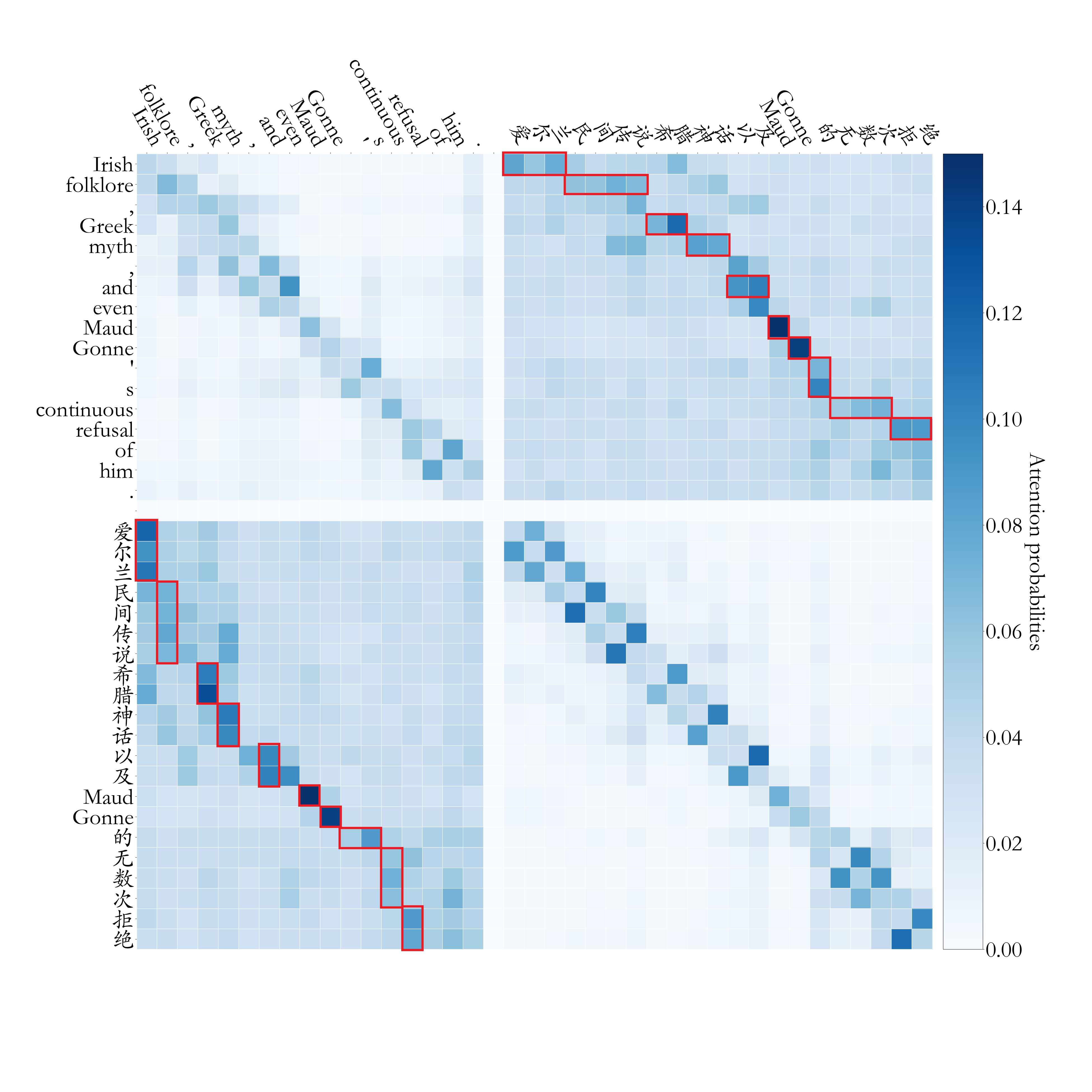}
		\caption{Decomposed attention.}
		\label{fig:attn_da_entity}
	\end{subfigure}
	\caption[l]{Alignment of entity, name \textit{Maud Gonne}.}
	\label{fig:attn_entity}
\end{figure}

\begin{figure}[!h]
	\centering
	\center 
	\begin{subfigure}[t]{.49\textwidth}
		\centering
		\includegraphics[width=1.\linewidth]{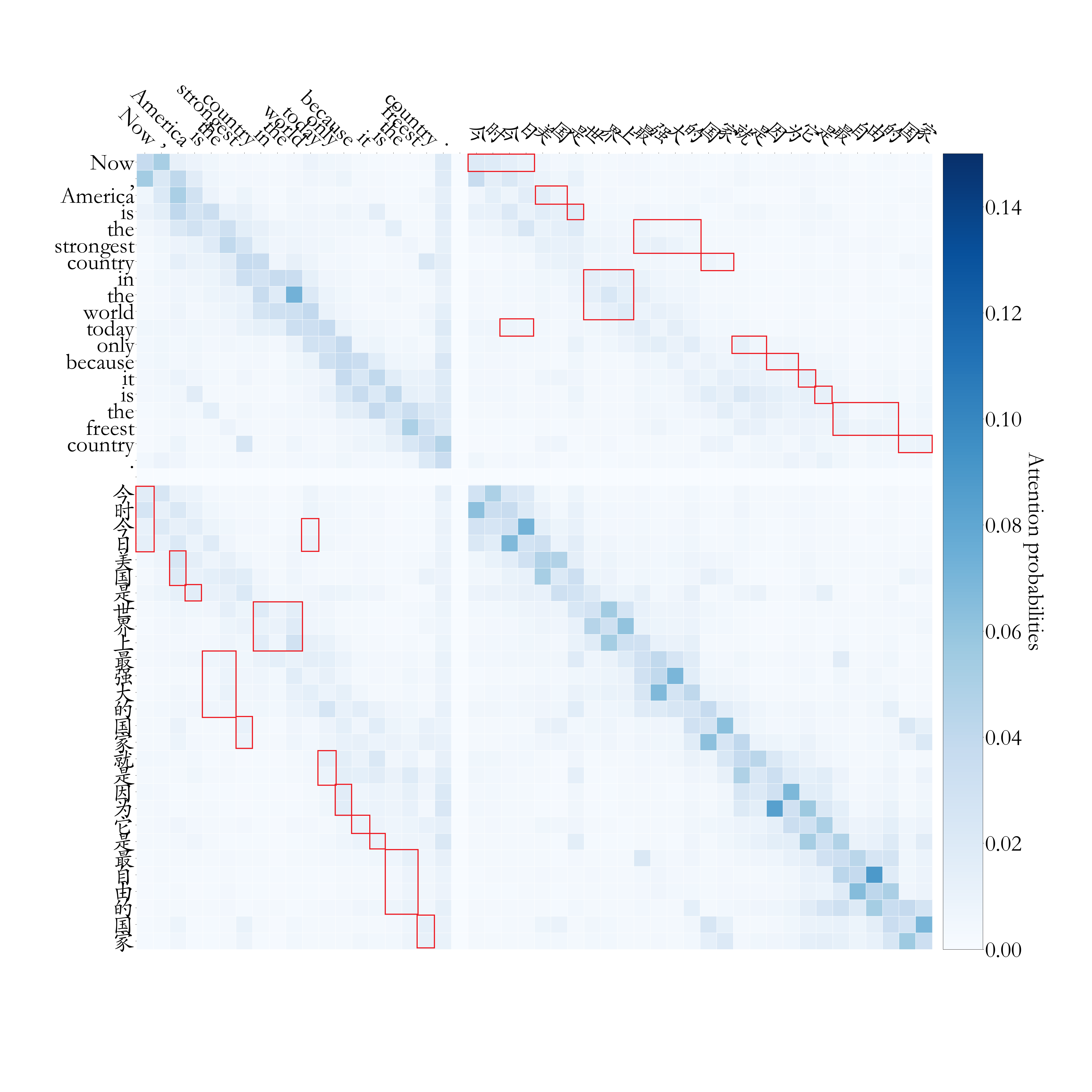}
		\caption{Mixed attention.}
		\label{fig:attn_ma_entity_country}
	\end{subfigure}
	\begin{subfigure}[t]{.49\textwidth}
		\centering
		\includegraphics[width=1.\linewidth]{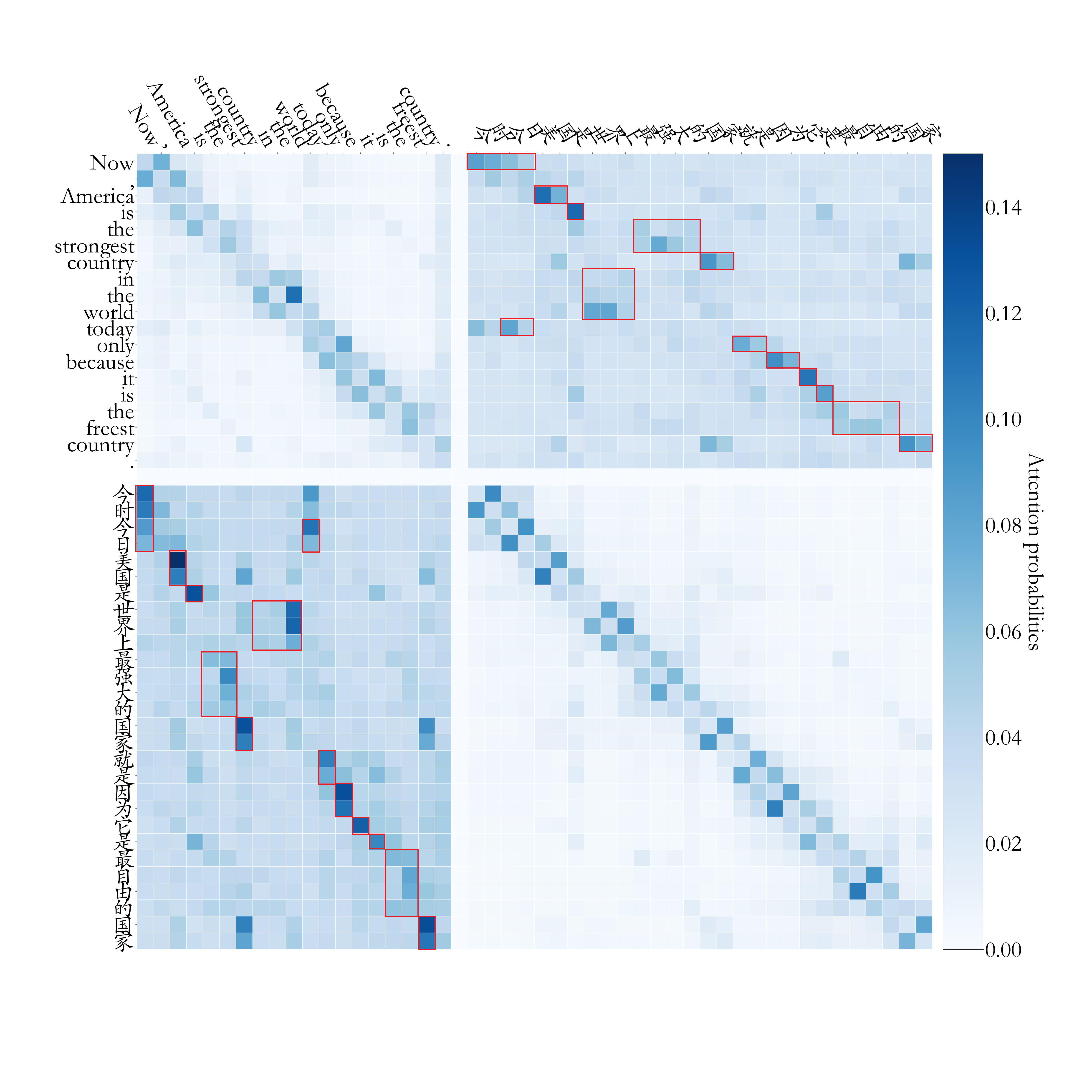}
		\caption{Decomposed attention.}
		\label{fig:attn_da_entity_country}
	\end{subfigure}
	\caption[l]{Alignment of entity, country \textit{America}.}
	\label{fig:attn_entity_country}
\end{figure}

\begin{figure}[!h]
	\centering
	\center 
	\begin{subfigure}[t]{.49\textwidth}
		\centering
		\includegraphics[width=1.\linewidth]{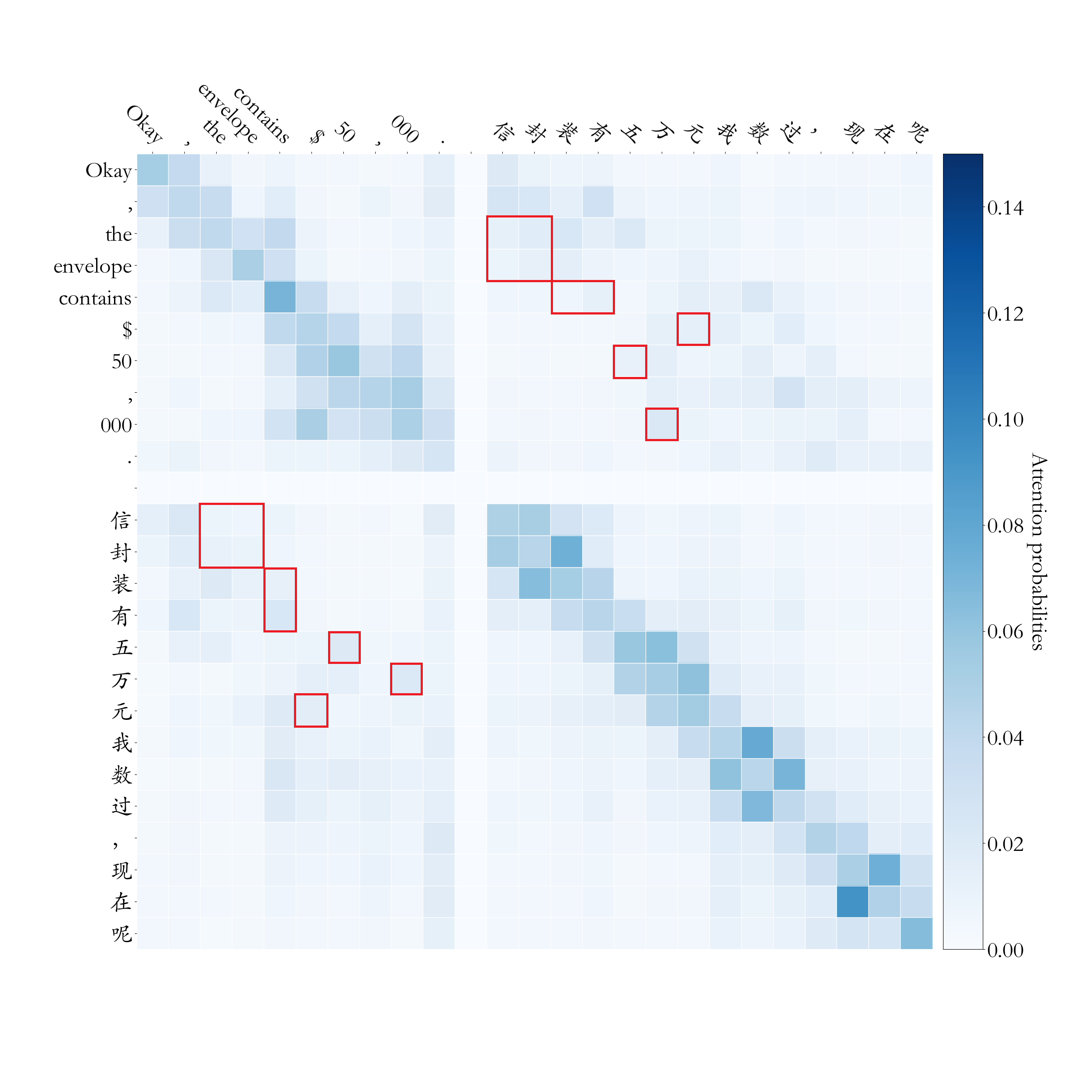}
		\caption{Mixed attention.}
		\label{fig:attn_ma_entity_currency}
	\end{subfigure}
	\begin{subfigure}[t]{.49\textwidth}
		\centering
		\includegraphics[width=1.\linewidth]{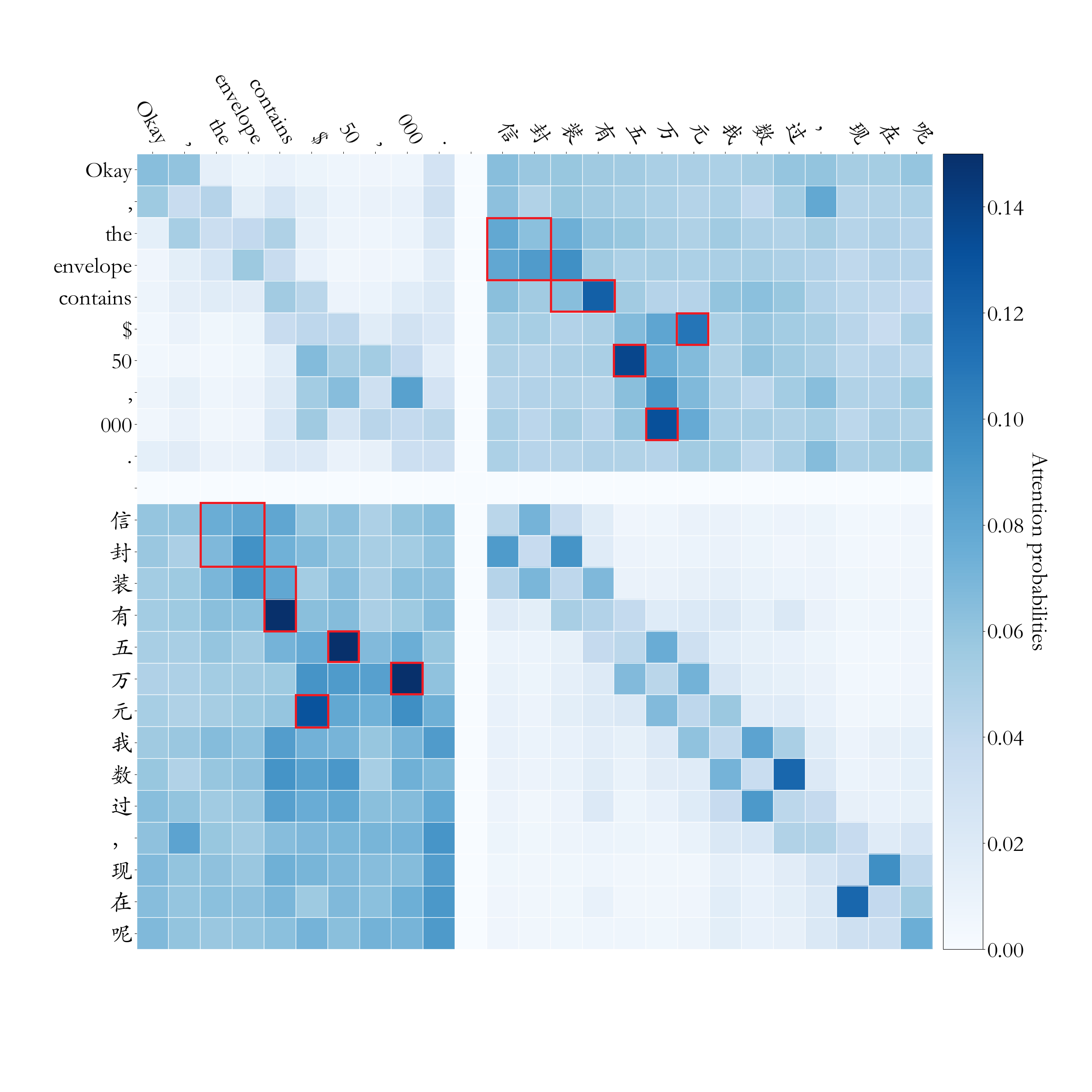}
		\caption{Decomposed attention.}
		\label{fig:attn_da_entity_currency}
	\end{subfigure}
	\caption[l]{Alignment of entity, currency \textit{\$50,000}.}
	\label{fig:attn_entity_currency}
\end{figure}

\begin{figure}[!h]
	\centering
	\center 
	\begin{subfigure}[t]{.49\textwidth}
		\centering
		\includegraphics[width=1.\linewidth]{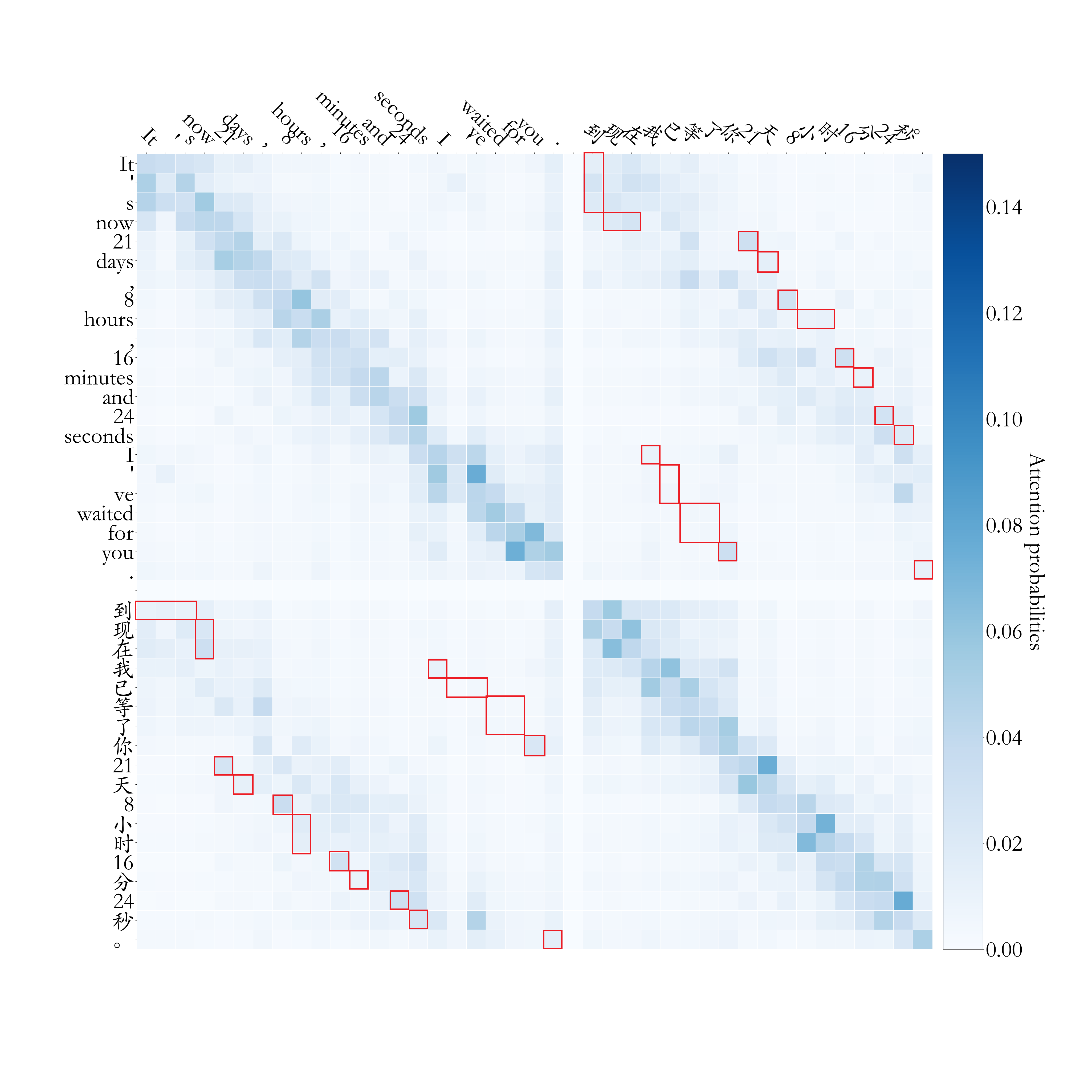}
		\caption{Mixed attention.}
		\label{fig:attn_ma_entity_time}
	\end{subfigure}
	\begin{subfigure}[t]{.49\textwidth}
		\centering
		\includegraphics[width=1.\linewidth]{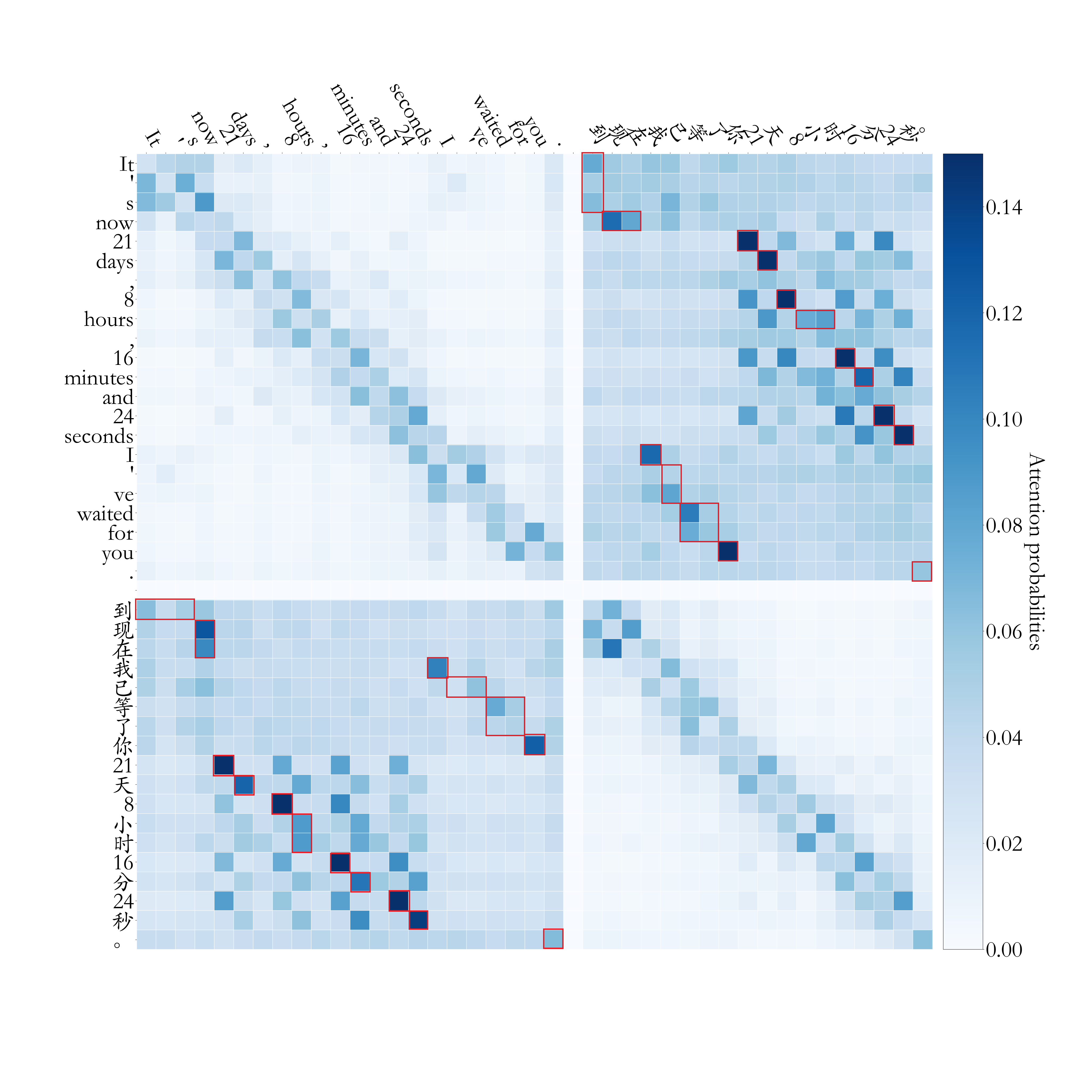}
		\caption{Decomposed attention.}
		\label{fig:attn_da_entity_time}
	\end{subfigure}
	\caption[l]{Alignment of entity, time \textit{21 days, 8 hours, 16 minutes and 24 seconds}.}
	\label{fig:attn_entity_time}
\end{figure}

\subsection{When Word Order is Different}

\begin{figure}[!h]
	\centering
	\center 
	\begin{subfigure}[t]{.49\textwidth}
		\centering
		\includegraphics[width=1.\linewidth]{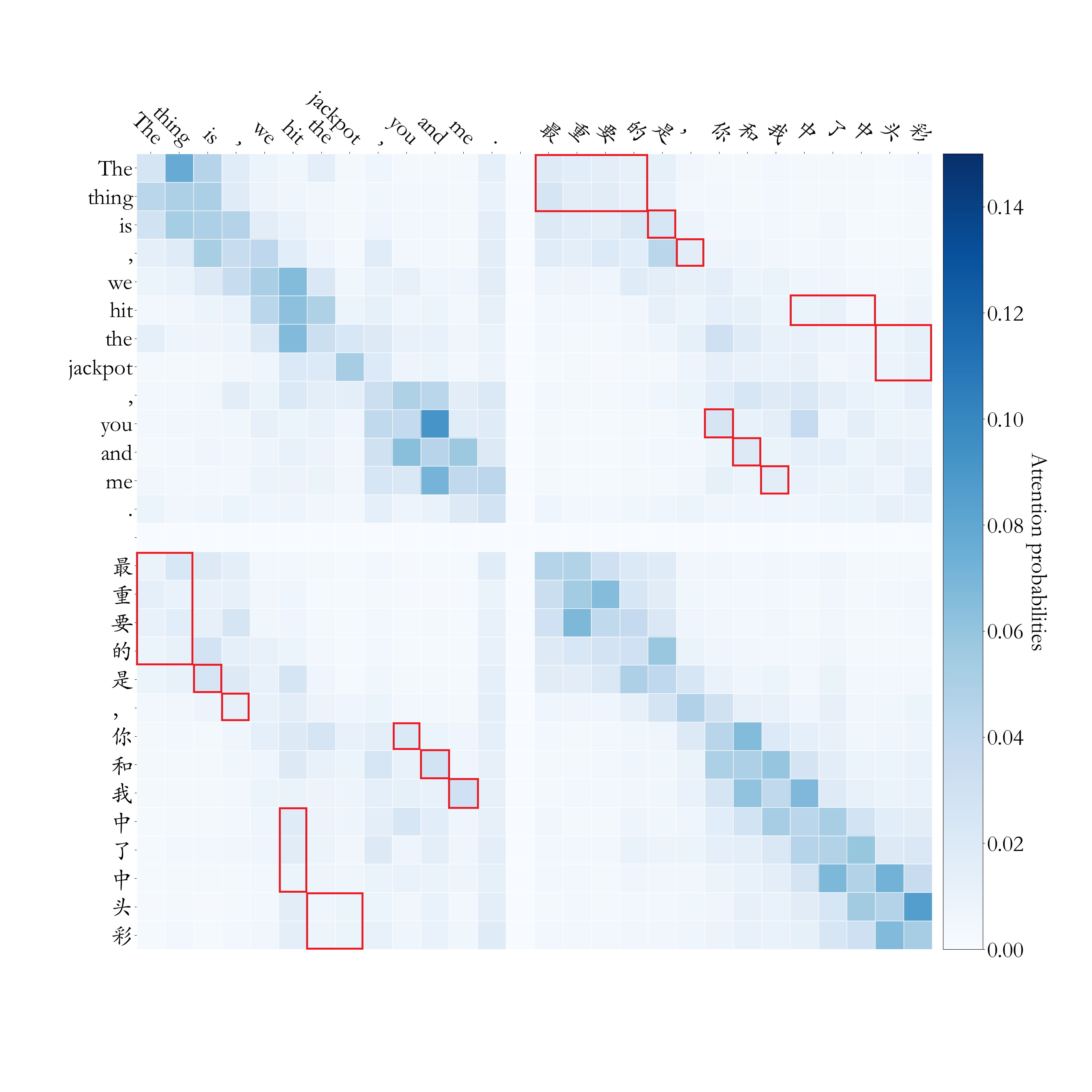}
		\caption{Mixed attention.}
	\end{subfigure}
	\begin{subfigure}[t]{.49\textwidth}
		\centering
		\includegraphics[width=1.\linewidth]{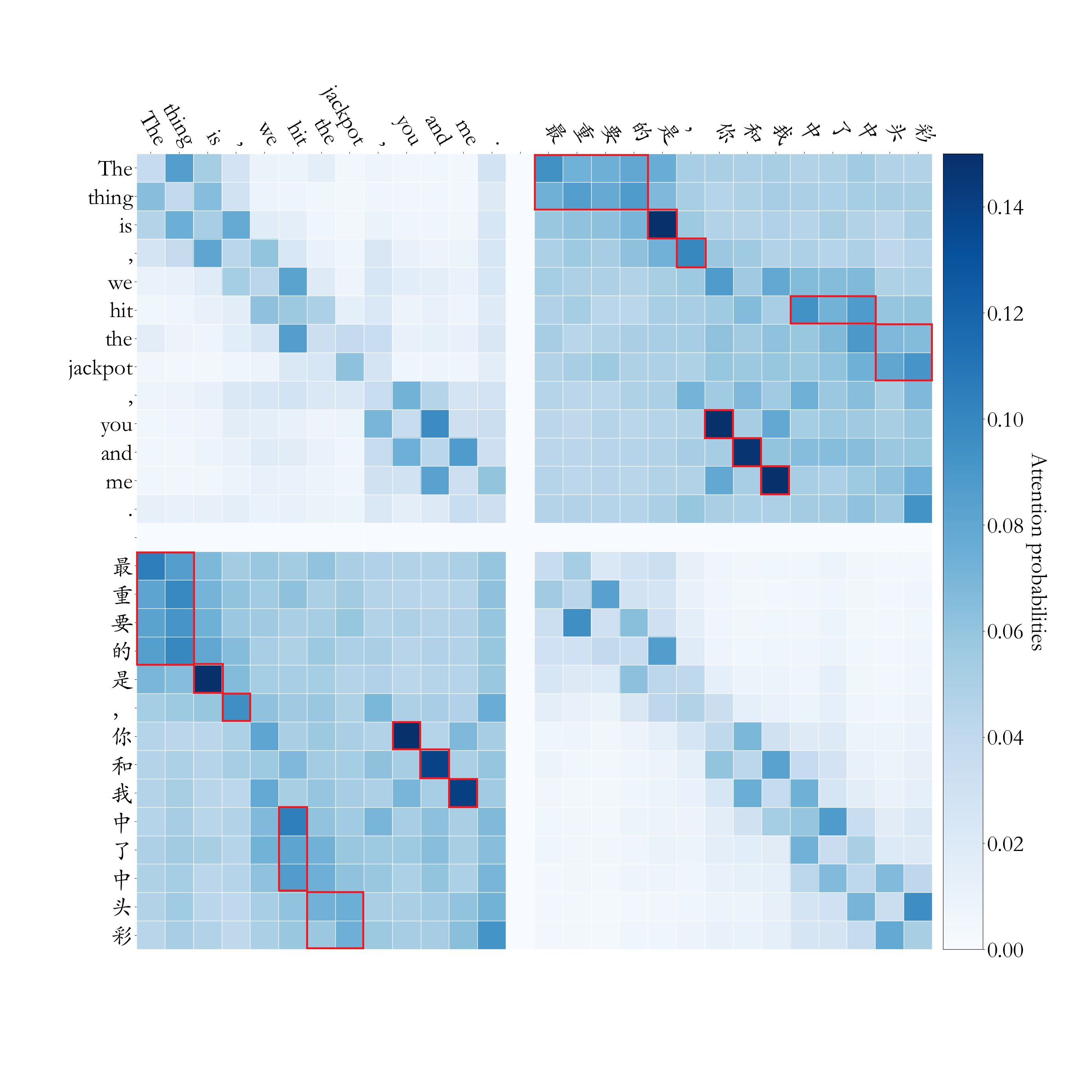}
		\caption{Decomposed attention.}
	\end{subfigure}
	\caption[l]{}
\end{figure}

\begin{figure}[!h]
	\centering
	\center 
	\begin{subfigure}[t]{.49\textwidth}
		\centering
		\includegraphics[width=1.\linewidth]{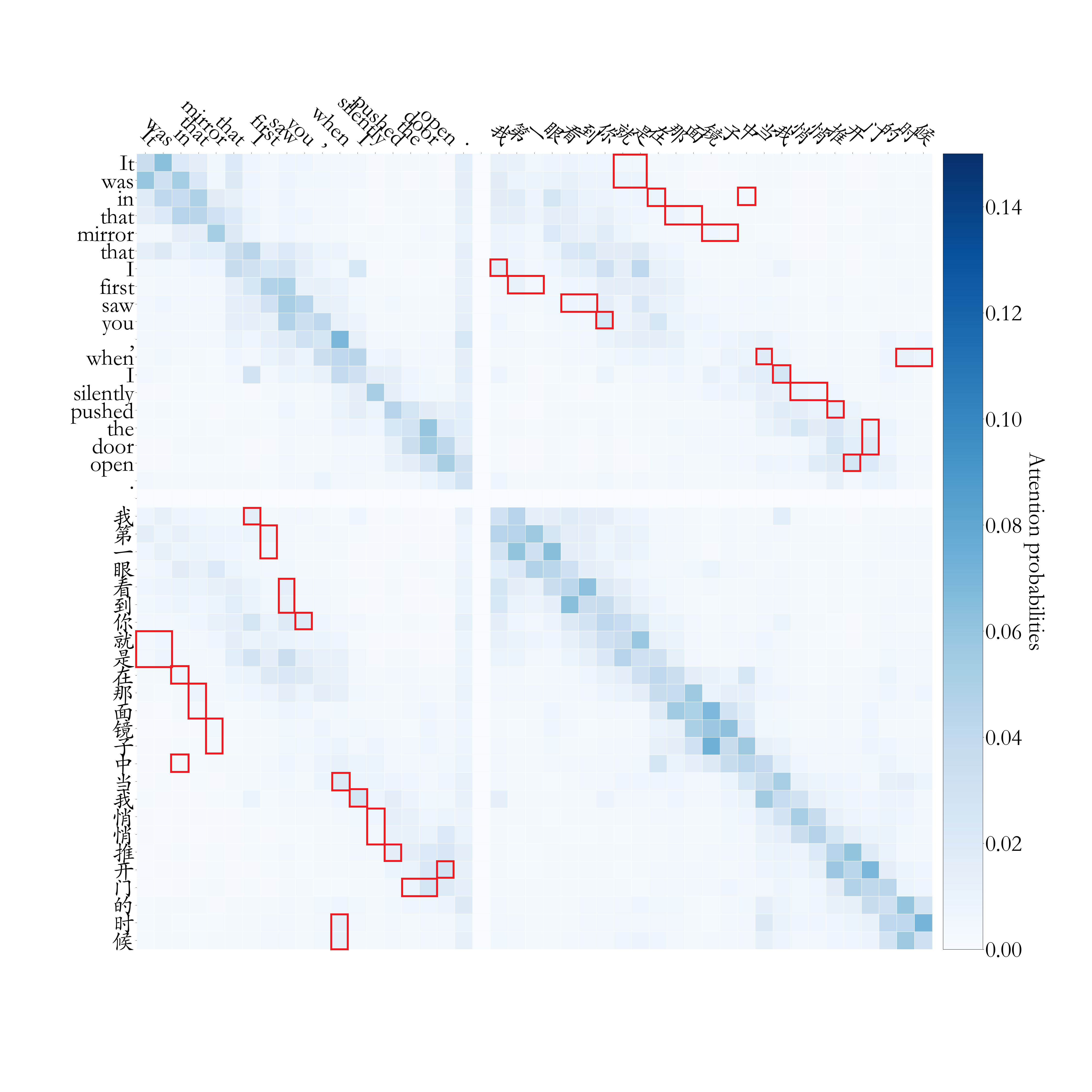}
		\caption{Mixed attention.}
	\end{subfigure}
	\begin{subfigure}[t]{.49\textwidth}
		\centering
		\includegraphics[width=1.\linewidth]{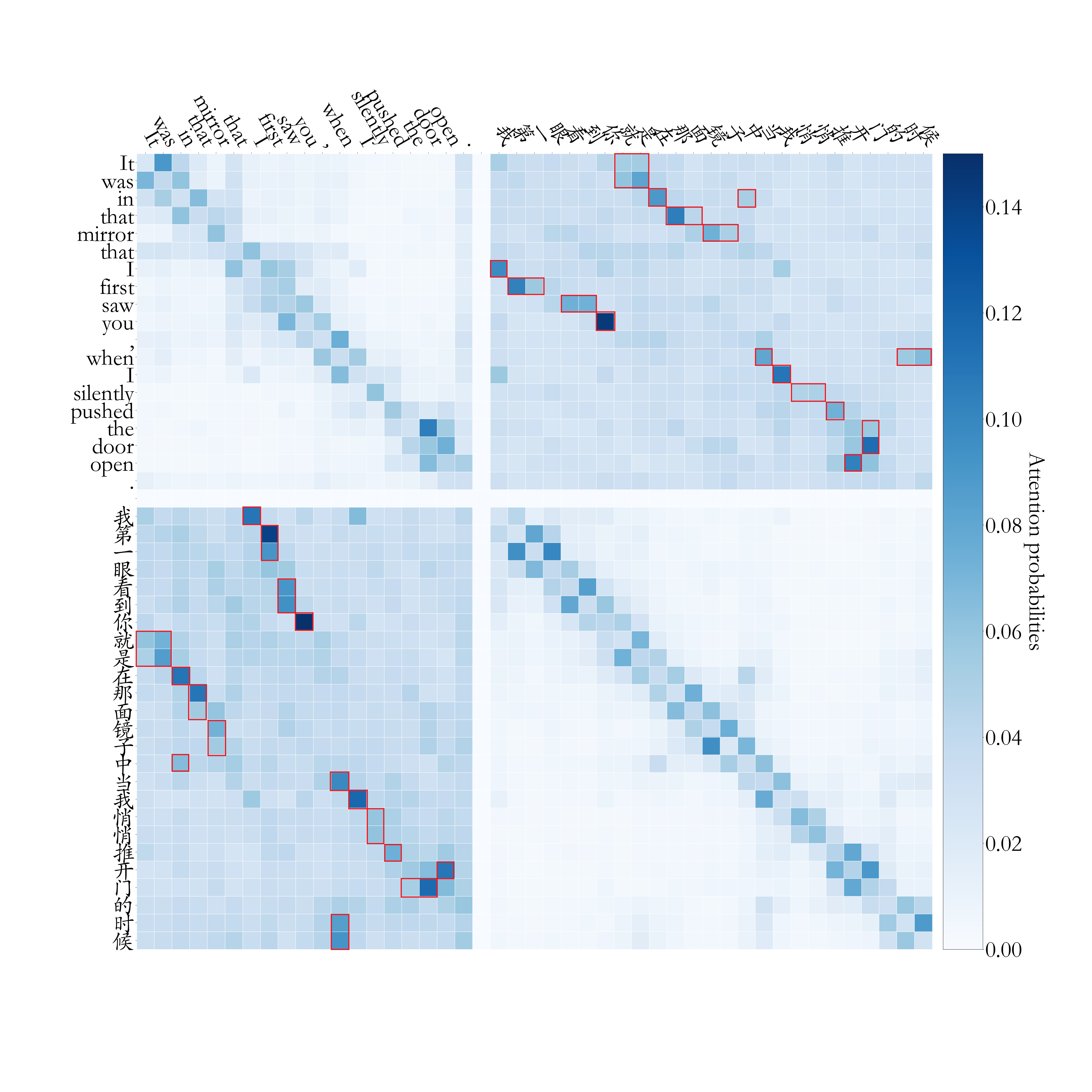}
		\caption{Decomposed attention.}
	\end{subfigure}
	\caption[l]{}
\end{figure}

\begin{figure}[!h]
	\centering
	\center 
	\begin{subfigure}[t]{.49\textwidth}
		\centering
		\includegraphics[width=1.\linewidth]{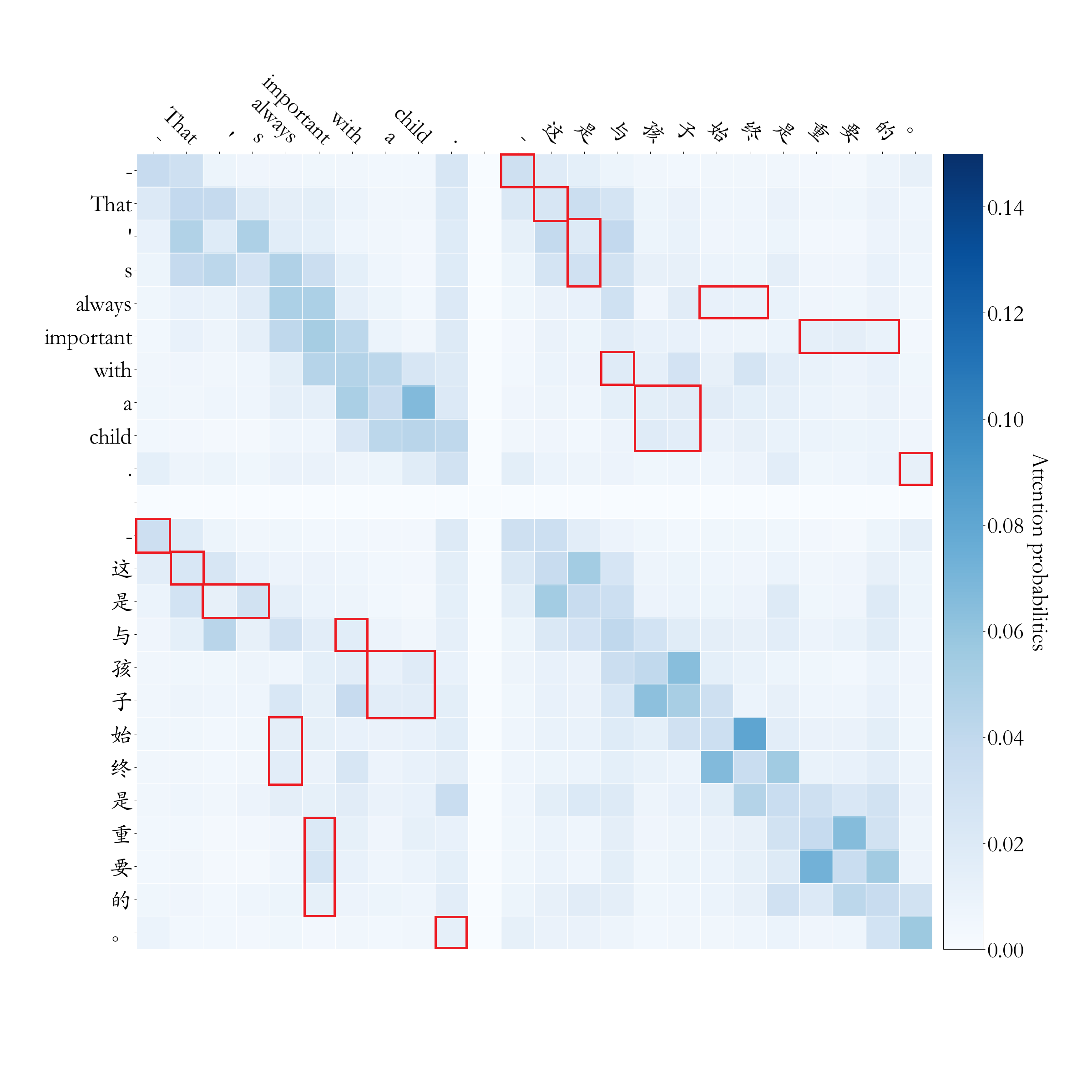}
		\caption{Mixed attention.}
	\end{subfigure}
	\begin{subfigure}[t]{.49\textwidth}
		\centering
		\includegraphics[width=1.\linewidth]{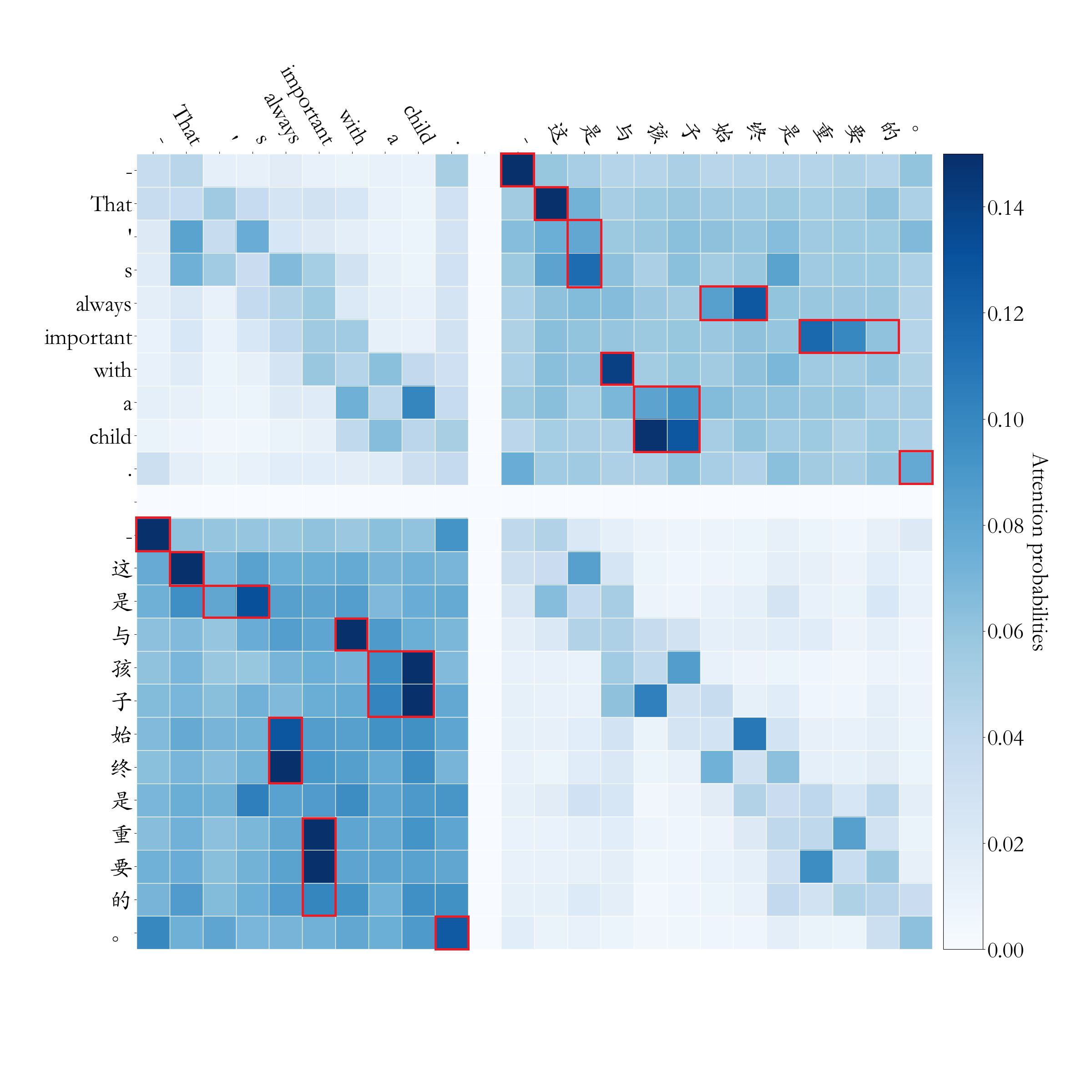}
		\caption{Decomposed attention.}
	\end{subfigure}
	\caption[l]{}
\end{figure}

\clearpage

\section{Multilingual Maksed Language Modeling}
\label{app:mmlm}

\begin{figure}[!h]
    \centering
	\includegraphics[width=.49\linewidth]{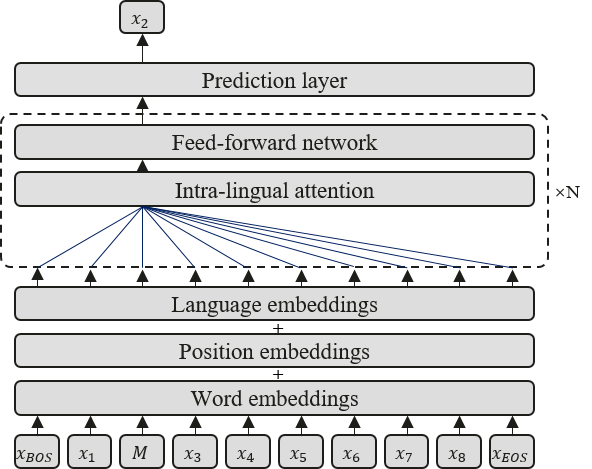}
	\caption{Multilingual masked language modeling (MMLM).}
	\label{fig:struct_mmlm}
\end{figure}

\section{Training Loss of Preliminary Experiment}
\label{app:lang_loss}

\begin{figure}[!h]
	\centering
	\includegraphics[width=.45\linewidth]{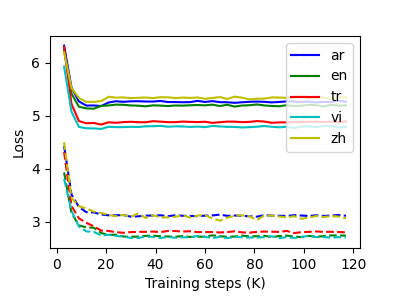}
	\caption[l]{Training CE loss of languages from different language families. \textit{Solid}: loss on monolingual data. \textit{Dashed}: loss on cross-lingual data. The experiment is conducted with MINI model as described in Section~\ref{sec:data_pt}.}
	\label{fig:lang_loss}
\end{figure}

\clearpage

\section{Hyper-Parameters}
\label{app:hyper}

\begin{table*}[!h]
    \centering
    \resizebox{0.6\textwidth}{27mm}{
    \begin{tabular}{c|c}
        \hline 
        Hyper-parameter & Value \\
        \hline 
        \hline 
        Hidden size & 768 \\
        \hline 
        FFN projection & 3072 \\
        \hline 
        Positional embedding length & 128 \\
        \hline 
        Warmup steps & 1800 \\
        \hline 
        Optimizer & LAMB~\citep{You2020Large} \\ 
        \hline 
        LAMB $\epsilon$ & 1e-4 \\
        \hline 
        LAMB $\beta_{1}$ & 0.9 \\
        \hline 
        LAMB $\beta_{2}$ & 0.999 \\
        \hline 
        Learning rate & 5e-5 \\
        \hline 
        Dropout probability & 0.1 \\ 
        \hline 
        Activation function & GELU~\citep{DBLP:journals/corr/HendrycksG16} \\ 
        \hline 
        Batch size & 2048 \\
        \hline 
        Attention layers & 12 \\
        \hline 
        Attention heads & 12 \\
        \hline 
        Training steps & 2011K \\
        \hline 
        Re-weighting $\mathcal{L}_T$ & 1.6 \\
        \hline 
        Re-weighting S & 2000K \\
        \hline 
    \end{tabular}
    }
    \caption{Hyper-parameters of pre-training BASE models.}
    \label{tab:hyper-params}
\end{table*}

\section{Dataset Statistics}
\label{app:dataset}

\begin{figure*}[h]
\centering
\includegraphics[width=1.\linewidth]{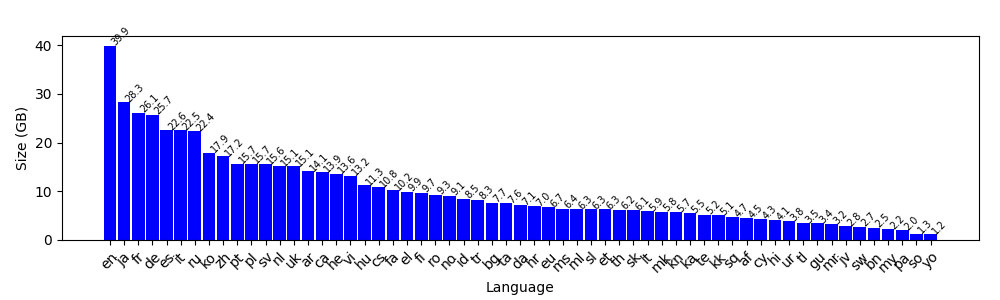}
\label{fig:data_wiki}
\caption{Statistics of Wikipedia corpus in our experiments.}
\end{figure*}

\clearpage

\begin{figure*}[h]
\centering
\includegraphics[width=1.\linewidth]{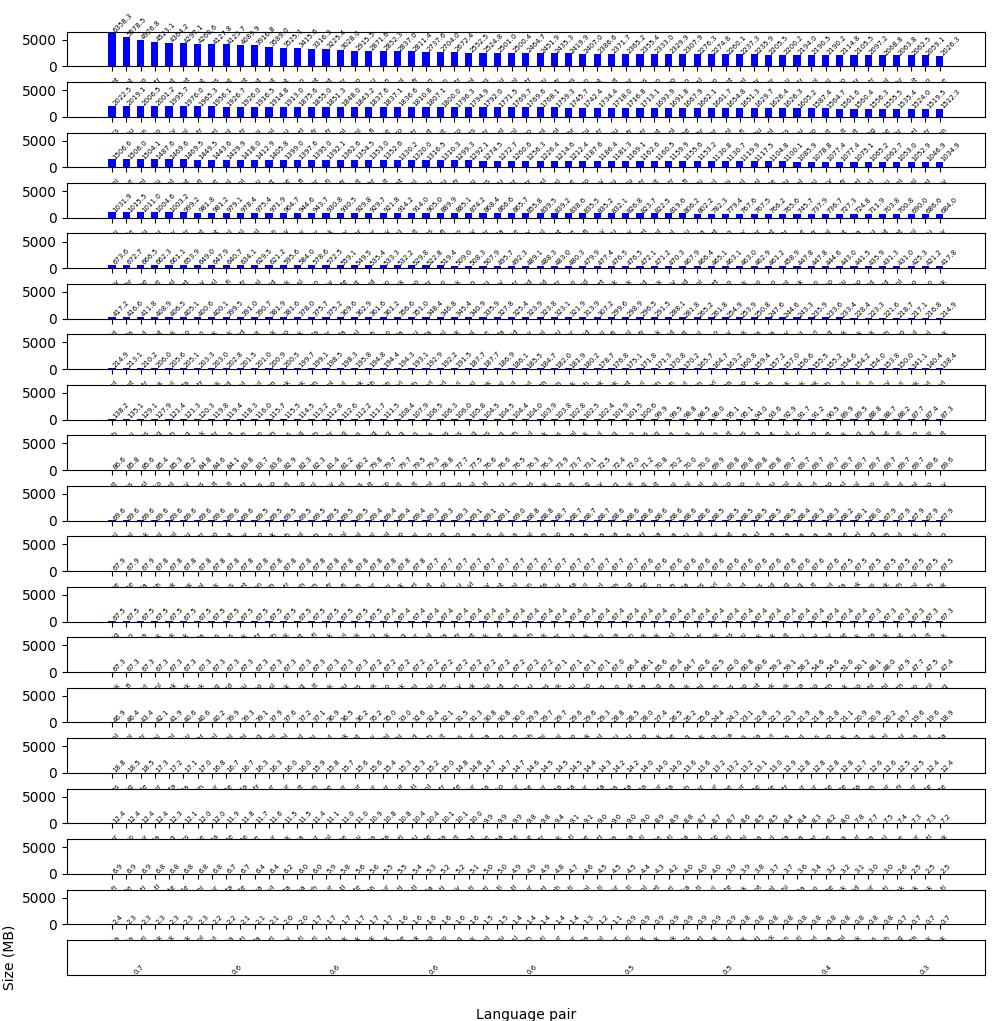}
\label{fig:data_opensubtitles}
\caption{Statistics of OpenSubtitles-2018 corpus in our experiments.}
\end{figure*}

\end{document}